\pdfoutput=1

\documentclass[11pt]{article}

\usepackage[preprint]{acl}
\usepackage{booktabs}
\usepackage{graphbox}
\usepackage{times}
\usepackage{latexsym}
\usepackage{amsmath}
\usepackage{bbm}
\usepackage{hyperref}
\usepackage{url}
\usepackage{algpseudocode}
\usepackage[linesnumbered,ruled]{algorithm2e}
\usepackage{algorithmicx}
\usepackage{amssymb}
\usepackage{pifont}
\usepackage{threeparttable}
\usepackage{multirow}

\usepackage[T1]{fontenc}
\RestyleAlgo{ruled}

\SetCommentSty{mycommfont}

\newcommand{\cmark}{\ding{51}}
\newcommand{\xmark}{\ding{55}}

\usepackage[utf8]{inputenc}
\usepackage{float}
\usepackage{arydshln}
\usepackage{microtype}
\usepackage{inconsolata}
\pdfoutput=1
\usepackage{graphicx}
\usepackage{svg}
\usepackage{hyperref}

\usepackage{amsmath}

\newcommand{\prompt}[3]
{
\begin{table}[th!]
\begin{minipage}{0em}
\fbox{\begin{minipage}{#1}
\centering
#2
\end{minipage}}
\fbox{\begin{minipage}{#1}
#3
\end{minipage}}
\end{minipage}
\end{table}
}

\newcommand{\promptLarge}[3]
{
\begin{table*}[th!]
\begin{minipage}{0em}
\fbox{\begin{minipage}{#1}
\centering
#2
\end{minipage}}
\fbox{\begin{minipage}{#1}
#3
\end{minipage}}
\end{minipage}
\end{table*}
}

\DeclareMathOperator{\mean}{mean} 
\DeclareMathOperator{\std}{std} 

\title{GRAF: Graph Retrieval Augmented by Facts for Romanian Legal Multi-Choice Question Answering}

\author{
\textbf{Cristian-George Craciun\textsuperscript{1,2}},
\textbf{Răzvan-Alexandru Smădu\textsuperscript{1}}, \\
\textbf{Dumitru-Clementin Cercel\textsuperscript{1}\thanks{Corresponding author.}},
\textbf{Mihaela-Claudia Cercel\textsuperscript{3,4}}
\\
\\
\textsuperscript{1}National University of Science and Technology POLITEHNICA Bucharest,\\
Faculty of Automatic Control and Computers, Bucharest, Romania\\
\textsuperscript{2}Technical University of Munich, Munich, Germany\\
\textsuperscript{3}Paris 1 Panthéon-Sorbonne University, Paris, France\\
\textsuperscript{4}University of Bucharest, Bucharest, Romania\\
 \small{cristian.craciun@tum.de, razvan.smadu@stud.acs.upb.ro, dumitru.cercel@upb.ro}
}

\begin{document}
\maketitle
\begin{abstract}
Pre-trained language models have shown remarkable performance in recent years, setting a new paradigm for natural language processing (NLP) research. The legal domain has received some attention from the NLP community, in part due to its textual nature. Question answering (QA) systems represent some of the tasks in this domain. This work explores the legal multiple-choice QA (MCQA) for Romanian. The contribution of this work is multi-fold. We introduce \textbf{JuRO}, the first openly available Romanian legal MCQA dataset, comprising 10,836 questions from three examinations. Along with this dataset, we introduce \textbf{CROL}, an organized corpus of laws comprising a total of 93 distinct documents with their modifications over 763 time spans, which we used for information retrieval techniques in this work. Additionally, we construct \textbf{Law-RoG}, the first graph of legal knowledge for the Romanian language, derived from the aforementioned corpus. Lastly, we propose a novel approach for MCQA, namely Graph Retrieval Augmented by Facts (\textbf{GRAF}), which achieves competitive results with generally accepted state-of-the-art methods and even exceeds them in most settings.
\end{abstract}

\begin{figure}[!th]
  \centering
  \includegraphics[width=\columnwidth]{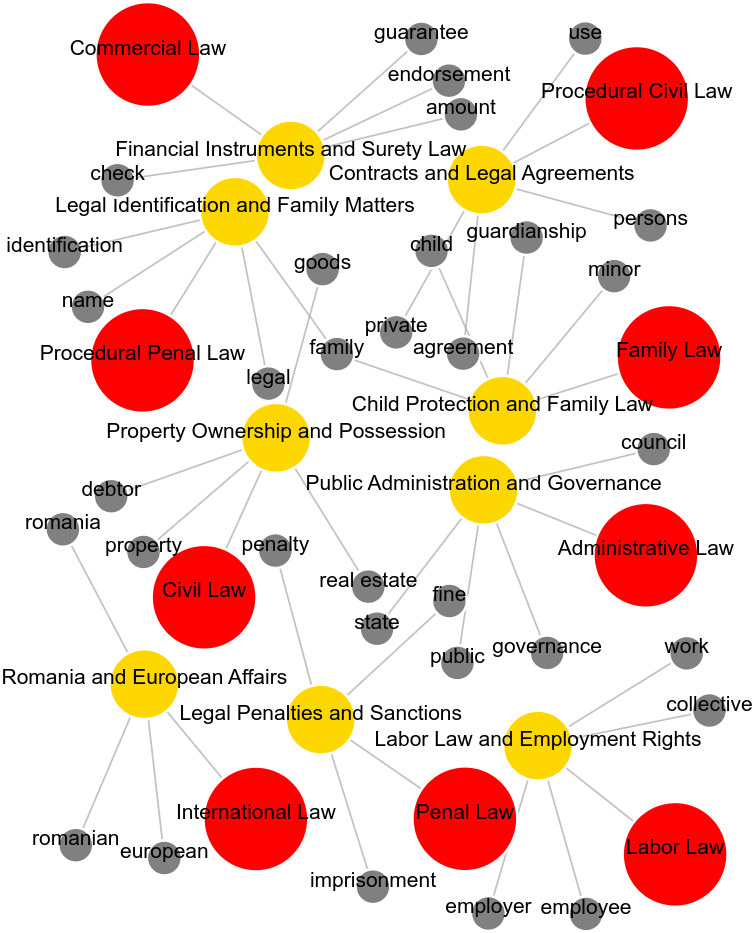}
  \caption{General topic graph for \textbf{CROL}. Red entities encompass the main legal branches, yellow entities represent topics, and grey entities are associated keywords.}
  \label{fig:banner}
\end{figure}

\section{Introduction}

Question answering (QA) represents a family of downstream natural language processing (NLP) tasks explored in various settings \citep{zhong2020jec, rogers2023qa, singhal2023towards}. Some QA tasks can be formulated as open-ended questions that require an elaborate answer, which may include the rationale behind it \citep{chen2017reading, labrak2022frenchmedmcqa}. Other tasks focus on retrieving the answer from a given context. In this scenario, the answer can either be explicitly found in the provided context, in which case a model has to identify the location at which the answer occurs \citep{zaib2021bert}, or infer the correct answer based on understanding and interpretation of the context at hand, a task known as machine reading comprehension \citep{baradaran2022survey}. 

The tasks mentioned above have been explored for various domains and languages \citep{hoppe2021towards, louis2021statutory, askari2022expert, sen2022mintaka, ekram2022banglarqa}. Our work focuses on multiple-choice question answering (MCQA) for the Romanian legal domain. Legal QA represents an emerging area of research due to the insights it can provide in addressing various other problems, such as the dialogue system \cite{he2024agentscourt}. From a practical perspective, the average citizen of a nation can benefit from the opportunity to find answers to legal inquiries alongside experts in the field, which can lead to increased productivity.


Additionally, natural language corpora for the Romanian language are scarce. A dataset for single-choice QA was recently performed by \citet{dima2024roqllama}, which proposed a Romanian medical MCQA dataset.
Although the Romanian legal domain has been tackled in the past \citep{masala2021jurbert, masala-etal-2024-improving}, no open-source datasets are available. This motivated us to introduce new comprehensive resources for the Romanian language.

In summary, the contributions of this paper are multifold:
\begin{itemize}
    \item We release \textbf{JuRO}\footnote{\label{ft:res}\url{https://github.com/craciuncg/GRAF}}, the first open-source legal dataset available for the Romanian MCQA.
    \item We release \textbf{CROL}\footnotemark[\value{footnote}], a structured corpus of Romanian law that can be utilized to query the law for answers (see Figure \ref{fig:banner}).
    \item We release \textbf{Law-RoG}\footnotemark[\value{footnote}], the first knowledge graph for the Romanian legal domain.
    \item We present a novel algorithm for the MCQA task called \textbf{GRAF}\footnotemark[\value{footnote}].
    \item We provide a comprehensive evaluation of our proposed MCQA dataset using state-of-the-art methods and our proposed method. We also released the code to allow future research to build on top of this work\footnotemark[\value{footnote}].
\end{itemize}

\section{Related Work}


\subsection{Legal-Domain Question Answering}

\noindent\textbf{English QA.}
Initially, traditional legal QA systems were based on information retrieval techniques, rule-based systems, support vector machines \cite{10.1007/978-3-319-10061-6_14,10.1007/978-3-319-61572-1_20}, and shallow convolutional networks (CNNs) \cite{kim2015convolutional} to answer legal questions from bar exams \citep{kim2015coliee}. Later, deep neural networks based on CNNs \cite{do2017legalquestionansweringusing} and BiLSTM \cite{john2017solving} were used with improved performance. The remarkable achievements of the Transformer models \citep{vaswani2017attention} motivated their use in this domain. \citet{shankar2023privacyglue} focused on privacy-related law, using various sources to build their benchmark dataset and evaluating several legal Bidirectional Encoder Representations from Transformers (BERT) models \citep{devlin2018bert}.
Similarly, \citet{DBLP:conf/nips/HendrycksBCB21} evaluated multiple BERT-based architectures on a newly proposed dataset, demonstrating that both the volume of training data and the architecture influence performance.
To support these findings, new datasets were proposed \cite{ravichander2019question,ahmadetal2020policyqa,sovrano2021dataset}, including large multidisciplinary datasets \cite{DBLP:conf/iclr/HendrycksBBZMSS21,chalkidis-etal-2022-lexglue,LegalBench}. \\


\noindent\textbf{Non-English QA.} Other languages also took the initiative in the early days. For example, the COLIEE shared task \cite{DBLP:journals/rss/RabeloGKKYS22}, proposed in 2014, was the first legal QA task on Japanese legal documents provided in both Japanese and English. \citet{bach2017vietnamese} proposed an analysis of the Vietnamese transportation laws and evaluated a CRF-based system as a prerequisite stage for QA. An encoder-decoder architecture was proposed by \citet{kien-etal-2020-answering}, which contained word embeddings, convolutional, and attention layers. The model was evaluated on a dataset containing 6,000 Vietnamese legal questions, outperforming existing retrieval-based methods. Later, \citet{vuong2023improving} proposed an end-to-end retrieval-based system that used a pre-trained BERT model on weakly labeled data. \citet{zhong2020jec} addressed the Chinese legal field, featuring an MCQA dataset containing practice exercise questions and a knowledge database. The comprehensive evaluation of various methods, including transformers, attention, and distant supervision, revealed a significant gap until human-level performance was achieved. Other works addressed languages such as Arabic \cite{hijazi-etal-2024-arablegaleval}, Chinese \cite{chen2023equals,jiang2024h}, French \cite{louis2021statutory,louis2024interpretable}, German \cite{buttner-habernal-2024-answering}, and Spanish \cite{calleja2021bilingual}.

\subsection{Romanian Legal Domain}

Recent advances in the Romanian legal domain have focused on developing specialized models, datasets, and tools. \citet{puaiș2021named} developed the LegalNERo dataset, a word embedding model, and a BiLSTM-CFR model for the legal named entity recognition (NER) task. \citet{smuadu2022legal} proposed a multi-task domain adaptation model to address legal NER. \citet{masala2021jurbert} proposed jurBERT, a BERT model adapted to the Romanian legal domain, trained on an extensive corpus of 160GB of legal text, achieving improved performance compared to RoBERT \cite{masala-etal-2020-robert}. Following this result, the authors attempted to build a judicial prediction system using this model \citep{masala-etal-2024-improving}. Other works focus on text classification~\cite{avram2021pyeurovoc} and anonymization of Romanian jurisprudence~\cite{puaics2024system}.

\subsection{Knowledge Graph-Based Question Answering}

The literature presents various approaches to knowledge graph-based QA. We identify two main classes of approaches: symbolic and numeric. Since a knowledge graph (KG) is a symbolic, structured representation of factual knowledge, some works revolved around symbolic methods. \citet{chakraborty2024multi} explored multi-hop QA approaches employing large language models (LLMs) and a KG-based algorithm, which identified the correct answer. The numeric approaches attempt to employ learned numeric representations of the data to make predictions. One such work is performed by \citet{he-etal-2022-dialmed}, which proposes a medical dialogue dataset as well as a method for utilizing a medical graph \citep{wang2019kgat} to predict the corresponding medication based on patient-doctor dialogue. Our work aims to create synergy between the two approaches and proposes new datasets and a knowledge graph for future use.

\section{Novel Resources}

\begin{table*}[t]
  \centering
  \small
  \begin{tabular}{lccccc}
    \toprule
    \textbf{Dataset} & \textbf{\# Examples} & \textbf{QA Format} & \textbf{Language} & \textbf{Public} \\
    \midrule
    PrivacyQA \citep{ravichander2019question} & 1,750 & Span & English & \cmark \\
    PolicyQA \citep{ahmadetal2020policyqa} & 714 & Span & English & \cmark \\
    JEC-QA \citep{zhong2020jec} & 26,367 & Multi-Choice & Chinese & \cmark \\
    COLIEE-21 \citep{rabelo2022overview} & 887 & Binary & Japanese & \cmark \\
    BSARD \citep{louis2021statutory}& 1,100 & Article Retrieval & French & \cmark \\
    EQUALS \citep{chen2023equals} & 6,914 & Long Form & Chinese & \cmark \\
    LLeQA \citep{louis2024interpretable} & 1,868 & Long Form & French & \cmark \\
    \noalign{\vskip 0.4ex}
    \multirow{2}{*}{LegalCQA \citep{jiang2024h}} & 21,780 & \multirow{2}{*}{Long Form} & Chinese & \multirow{2}{*}{\cmark} \\
    & 8,899 & & English & \\
    \hdashline\noalign{\vskip 0.7ex}
    JuRO (Ours) & 10,836 & Multi-Choice & Romanian & \cmark \\
    \bottomrule
  \end{tabular}
  \caption{Comparison of JuRO with other existing datasets.}
  \label{tab:juro_comp}
\end{table*}

\begin{table*}[t]
  \centering
  \small
  \begin{threeparttable}
  \begin{tabular}{lccccc}
    \toprule
    \textbf{Corpus} & \textbf{Size} & \textbf{Dataset Type} & \textbf{Language} & \textbf{Public}\\
    \noalign{\vskip 0.4ex}
    \hline\noalign{\vskip 0.4ex}
    Marcell \citep{varadi2020marcell} & 317k documents, 774M tokens & Pre-training & Multilingual\tnote{*} & \cmark \\
    RoJur \citep{masala2021jurbert} & 11M entries, 160GB & Cases & Romanian & \xmark \\
    \citet{10.1145/3383583.3398616} & 200k court rulings & Rulings & German & \cmark \\
    \citet{collarana2018question} & 62 pages, 64 sections, 24k words & Span & English & \xmark \\
    \citet{kien-etal-2020-answering} & 8.5k documents, 118k articles & Article & Vietnamese & \xmark \\
    EQUALS \citep{chen2023equals} & 3,081 articles & Article & Chinese & \cmark \\
    \hdashline\noalign{\vskip 0.7ex}
    CROL (Ours) & 330k articles, 31.5M words & Article & Romanian & \cmark \\
    \bottomrule
  \end{tabular}
  \begin{tablenotes}\footnotesize
  \item[*] Bulgarian, Croatian, Hungarian, Polish, Romanian, Slovak, Slovenian 
  \end{tablenotes}
 \end{threeparttable}
  \caption{Comparison of CROL with other datasets from various works.}
  \label{tab:crol_comp}
\end{table*}

\subsection{JuRO}

We introduce a new dataset for legal QA, called JuRO. It contains past examination questions from all the legal branches examined in Romania. It represents the first dataset of its kind that we release to the public (see also Table \ref{tab:juro_comp}). \\
 
\noindent\textbf{Dataset Construction.} 
The data is extracted from various official examination portals.
Some of the subjects were extracted using OCR. To avoid errors, we manually inspected all the data, and thus, the resulting samples present minimal damage. Each entry essentially consists of a body in which a theoretical question is posed regarding a legal aspect, along with three possible answer choices labeled A, B, and C, out of which at most two answers are correct. \\

\noindent\textbf{Statistics.}
The dataset contains questions from three types of examinations: entrance into the judicial system (i.e., entrance), entrance into the bar (i.e., bar), and promotion exams for judicial positions (i.e., promotion). We present the distribution among legal domains and possible answer combinations in Figure \ref{fig:juropie} of the Appendix \ref{appendix:analysis}.
Promotion exams have three possible choices with a single correct answer, whereas the others have up to two possible correct answers. The distribution among correct answers is generally balanced, with a small exception for bar exams. However, it should be noted that the questions with a single correct answer are predominant and balanced. Data analysis for the JuRO dataset is presented in Appendices \ref{appendix:analysis} and \ref{appendix:topic_analysis}. \\

\noindent\textbf{JuRO vs. Existing Work.}
We introduce a new dataset for the Romanian legal QA, which encompasses three different types of examinations: entrance, bar, and promotion. We are the first to propose a legal MCQA dataset for the Romanian language and to make it publicly available. We hope that this will open opportunities for future research in Romanian, multilingual, and low-resource language settings. In Table \ref{tab:juro_comp}, we compare our work with other legal datasets. Although it is less than half the size of the JEC-QA dataset \cite{zhong2020jec}, it is larger than other existing datasets for legal QA in other languages.

\subsection{CROL}

\begin{table}[th]
  \centering
  \resizebox{\columnwidth}{!}{
  \begin{tabular}{lccccc}
    \toprule
     & 
    \textbf{Count} & \textbf{Avg. Length} & \textbf{Max Length} \\
    \midrule
    Articles & 330,320 & 95.11 & 24,735 \\
    Words & 31,416,577 & 6.16 & 28\\
    Vocabulary & 78,355 & 9.59 & 28\\
    \hdashline\noalign{\vskip 0.7ex}
    Nodes & 160,402 & - & - \\
    Edges & 319,958 & - & - \\
    \bottomrule
  \end{tabular}
  }
  \caption{General statistics for CROL and Law-RoG KG.}
  \label{tab:corpstats}
\end{table}

\textbf{C}orpus for \textbf{Ro}manian \textbf{L}aw (CROL) represents a collection of legal documents collected for law branches as follows: \textit{civil}, \textit{penal}, \textit{work}, \textit{administration}, \textit{commercial}, \textit{family}, and \textit{international}. \\

\noindent\textbf{Dataset Construction.} The CROL corpus was constructed using the official Ministry of Justice department portal\footnote{\url{https://legislatie.just.ro/}} to crawl all laws from the covered branches in the JuRO dataset. All these resources have been extracted from official sources of national state institutions. Therefore, the language is formal and presents few or no grammatical errors. \\

\noindent\textbf{Statistics.} CROL represents a collection of organized legal documents, corresponding to 93 distinct laws and 768 different versions of these laws. It contains 330k articles totaling almost 31.5M words with a vocabulary of 78.3k words. Statistics are also presented in Table \ref{tab:corpstats} and Appendix~\ref{appendix:analysis} and \ref{appendix:topic_analysis}. See also Figure \ref{fig:banner} for a graphical view of the topics and keywords in the corpus. \\

\noindent\textbf{CROL vs. Existing Work.} 
Our corpus can mainly serve as a knowledge base for information retrieval (IR) techniques for Romanian legal tasks. There has been past work on creating a Romanian legal corpus, such as the Marcell project \citep{varadi2020marcell}, which aimed to develop a multilingual legal corpus that includes Romanian law. However, this represents an annotated text corpus useful for NER training in a legal context, as well as related, but it does not make a distinction between
documents that are in effect and those that are not. We have made a clear separation between legal documents and their updates. There are also other efforts \cite{collarana2018question,kien-etal-2020-answering,masala2021jurbert}; however, they are not publicly available (see Table \ref{tab:crol_comp}).

\subsection{Law-RoG}

We introduce the first knowledge graph for the Romanian language, namely Law-RoG. This KG is built on the CROL corpus via entity-relation extraction. In particular, following the work of \citet{edge2024local}, we prompt an LLM to identify named entities and the relations between them to output entity-relation-entity triplets in our desired format using in-context learning \citep{brown2020fewshot} abilities that LLMs exhibit via few shot prompting (see Appendices \ref{appendix:ro_prompts} and \ref{appendix:en_prompts}). We opted for this approach because the Romanian NLP lacks resources for building specialized pipelines for entity and relation extraction, particularly in the legal domain. To validate that the LLM produced factually and coherently correct information, we asked 5 human NLP experts to evaluate 10 randomly sampled documents and their corresponding generated relations for each legal domain. They all agreed that the outputs were coherent and did not hallucinate beyond the given document. Although not perfect, we concluded that the generated relations were appropriate for almost all related NLP applications. The resulting KG spans 160k nodes and 320k edges (see Table~\ref{tab:corpstats}).

\section{GRAF}

We introduce a novel approach for retrieving information from a KG, which we applied to the legal MCQA. The same principles discussed regarding claim checking and validation can be applied to other tasks requiring factual knowledge.

\subsection{Problem Formulation}\label{sec:problem}

At first glance, we are presented with questions that may have only one correct answer for a dataset, and in other scenarios, a question can have up to two correct choices. We will make a distinction between these two settings and formulate the goal of the problem according to the architecture of the proposed model.

The multi-choice QA problem can be formulated as follows. Consider a dataset \(\mathcal{D}=\{x_i = (Q_i, C_i, T_i) \},i=1:\left|\mathcal{D}\right|\), where each triplet entry \(x_i\) contains the question body \(Q_i\) in textual form, the set of $|C_i|$ possible answer choices \(C_i = \{C^k_i\}, {k=1:|C_i|}\) and the set of target answers \(T_i\) with \(|T_i| \leq |C_i| \) and \(T_i \subseteq C_i\). Each answer choice is a tuple \(C^j_i=(\sigma^j_i, \epsilon^j_i)\) where \(\sigma^j_i\) is the choice label (e.g., A, B, C) of the answer, and \(\epsilon^j_i\) represents its textual content. In our setting, we examine questions with $|C_i|=3$ choice answers and $|T_i| \in \{1,2\}$ correct answers, i.e., single-choice and multi-choice QA, depending on the dataset. Moreover, we investigate two classes of models designed to address these QA variations and formulate the learning goal in Appendix \ref{appendix:arch}.

\begin{figure*}[ht!]
  \includegraphics[width=\linewidth]{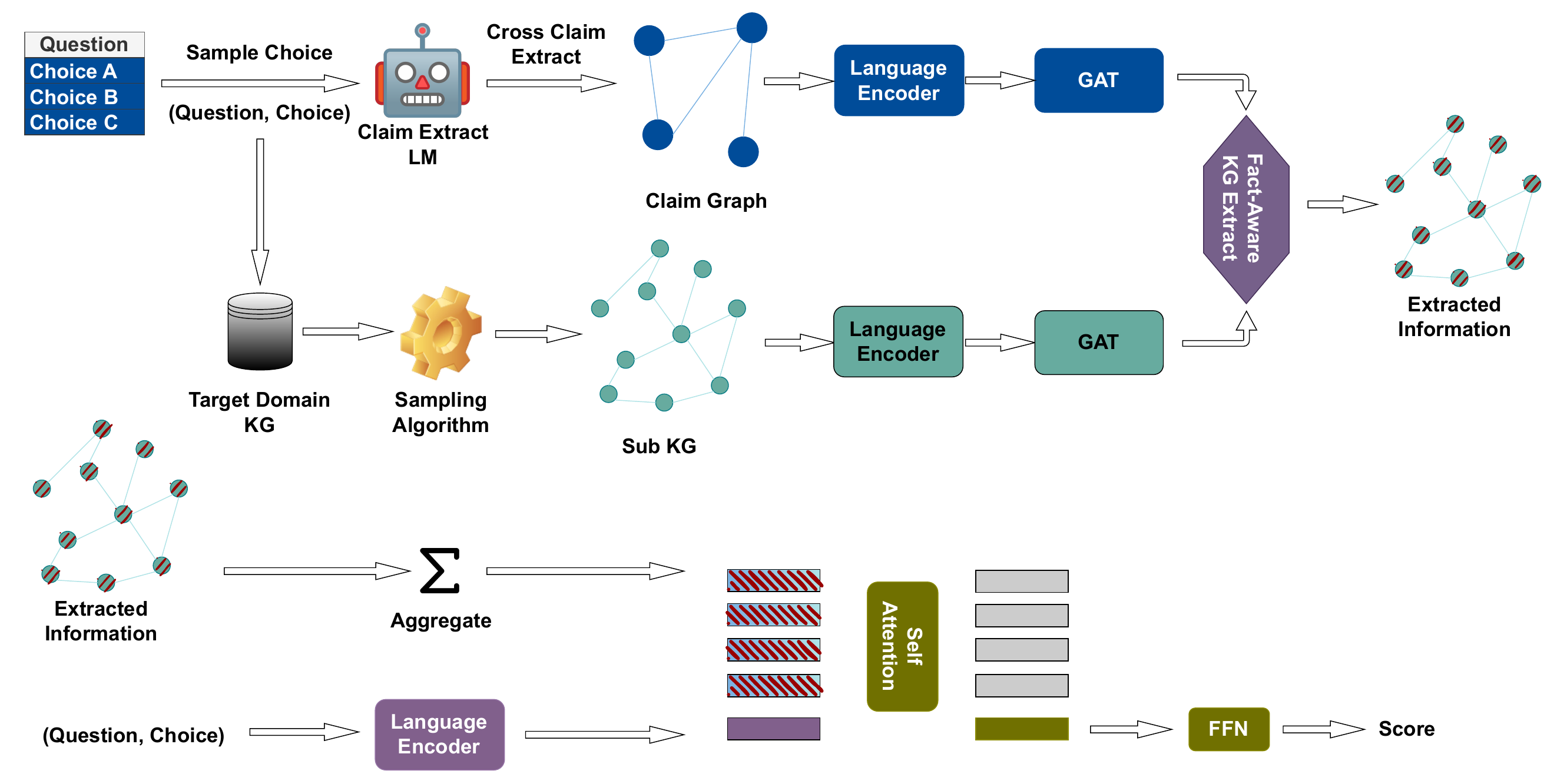}
  \caption {\textbf{GRAF} procedure overview.}
  \label{fig:graf_algo}
\end{figure*}

\begin{figure*}[t]
  \includegraphics[align=c, width=\linewidth]{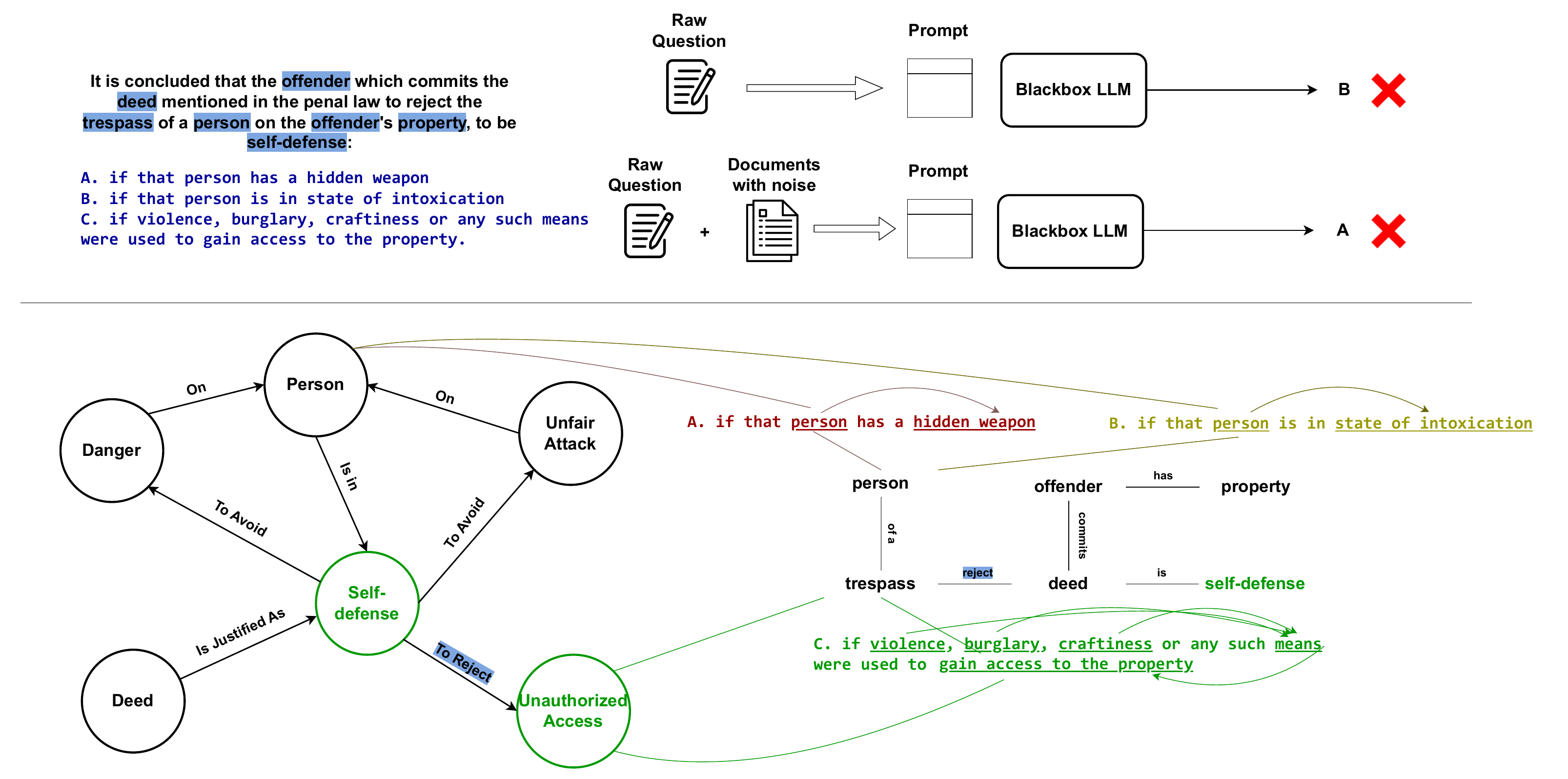}
  \caption{A showcase of how \textbf{GRAF} works on a sample question from the \textbf{JuRO} dataset translated to English. First, we construct the claim graph using an LLM that extracts the entities (nodes) and relations between them (edges). Based on the sub-KG extracted from Law-RoG and the claim graph, we determine that the entities ``Self-defense'' and ``Unauthorized Access'' have the best alignment with the entities in choice ``C'', thus, it is most likely to be the correct answer.}
  \label{fig:graf_sample}
\end{figure*}

\begin{algorithm}[th]
\caption{GRAF}\label{alg:two}
\small
\KwData{$Q$ - question, $C$ - choices, $G$ - knowledge graph}
\KwResult{$P$ - choice probabilities}
\(P \gets \text{[]}\)\\
\For{$c_i \in split\_choices(C)$}{
    \tcp{Query the cross-claim extraction model and obtain the claim graph CG}
    \(CG \gets \text{cross\_claim\_extract}(Q||c_i)\) \\
    \tcp{Sample a subgraph \(SG\) from \(G\)}
    \(SG \gets \text{sample\_graph}(G)\) \\
    \tcp{Encode \(SG\)'s components}
    \(SGE, CGE \gets \text{\{\}}, \text{\{\}}\)\\
    \For{$(E_a, r_{ab}, E_b) \in SG$}{
        \(SGE\text{.append}(\text{enc}(E_a), \text{enc}(r_{ab}), \text{enc}(E_b))\)
    }
    \tcp{Encode \(CG\)'s components}
    \For{$(E_a, r_{ab}, E_b) \in CG$}{
        \(CGE\text{.append}(\text{enc}(E_a), \text{enc}(r_{ab}), \text{enc}(E_b))\)
    }
    \tcp{Compute the alignment between encoded claims and all encoded relations}
    \For{$(h^i_c, h^j) \in \text{GAT}(SGE) \times \text{GAT}(CGE)$}{
        \(R^{ij} \gets \cos(h^i_c, h^j)\)\\
    }
    \tcp{Compute the embedding for all neighboring claims}
    \(\Bar{H} \gets Rh\)\\
    \tcp{Compute the final score using self-attention}
    \(c \gets \text{enc}(Q||C)\)\\
    \(c_{\text{final}}, H_{\text{final}} \gets \text{SelfAttention}(c||\Bar{H})\)\\
    \tcp{Save the score of the current choice}
    \(P\text{.append}(c_{\text{final}})\)
}
\end{algorithm}

\subsection{Algorithm Description}

Our proposed algorithm is applied to every given question and each of its possible answer choices. Specifically, the inputs to GRAF are the question-choice pair and the KG, which contains entities as nodes and relationships between entities as edges. The algorithm is illustrated in Figure \ref{fig:graf_algo} and Algorithm \ref{alg:two} and will be presented in what follows. \\

\noindent\textbf{Claim Graph.} A multiple-choice question is primarily composed of choices that present different claims with various truth values. We are interested in determining claims whose entities come from (1) the question and the given choice or (2) both come from the choice. In our setting, questions generally present hypothetical scenarios or premises that do not contradict the law; therefore, we do not consider the question body itself to present any untrue claims. 
Therefore, to choose the most suitable answer, we decompose each pair (question, choice) into its underlying claims using an LLM, similar to \citet{edge2024local}; however, any lighter solutions can be adopted with adequate resources. We call this procedure \textit{Cross Claim Extraction}. An example is presented in Figure \ref{fig:graf_sample}. \\

\textbf{KG Sampling.} We obtain the domain-specific KG (in our setting, from Law-RoG) by using the law branch related to the question. Since it is often infeasible to consider the entire graph for inference due to its size, we resort to sampling the graph via a procedure that retrieves the most relevant nodes and edges using a heuristic. 
First, we preprocess the words by tokenizing and lemmatizing them using the \textit{SpaCy}\footnote{\url{https://spacy.io}} package for the Romanian language to achieve better, less noisy results. Then, we use a bag-of-words (BoW) approach and incorporate the BM25 retriever \citep{robertson1976relevance} to select the top k entities in the knowledge graph. We proceed to select their vicinity with a breadth-first search for a limited depth. We also limit the maximum number of entities retrieved during this stage. In our work, we use a depth of 1, select the top 10 entities, nodes, and edges from the KG, and limit the selection to no more than 50 distinct entities. \\

\noindent\textbf{KG Encoding.} To encode the KG, we utilize a language encoder model to embed the entities and relations, resulting in sets of node and edge features \(\boldsymbol{h_N}\) and \(\boldsymbol{h_E}\), respectively. Then, we adapt the Graph Attention Network (GAT) \citep{velickovic2018graph} to further capture the topological information for each entity. The original GAT model was developed only for graphs with no relation encoding.
Therefore, we transform the set of features using two different linear transformations, parametrized by shared \(\boldsymbol{W_N}\) and \(\boldsymbol{W_E}\) for the nodes and edges, respectively.
To capture the relational topological information, we compute the attention coefficients for the relations \(\boldsymbol{e^{ij}_E}\) in which the current entity is involved between the current node \(i\) and the adjacent edges \(j\):

\begin{equation}
    \label{eq:relatt}
    \boldsymbol{e^{ij}_E} = \sigma_A(\boldsymbol{(a_E)^T} [\boldsymbol{W_Nh^i_N} || \boldsymbol{W_Eh^j_E}])
\end{equation}
and calculate the attention coefficients for the nodes \(e^{ij}_N\) to capture inter-entity relations between the current node \(i\) and neighboring nodes \(j\):

\begin{equation}
    \label{eq:nodeatt}
    \boldsymbol{e^{ij}_N} = \sigma_A(\boldsymbol{(a_N)^T} [\boldsymbol{W_Nh^i_N} || \boldsymbol{W_Nh^j_N}])
\end{equation}
where \(\boldsymbol{a_N}\) and \(\boldsymbol{a_E}\) represent two distinct attention vectors for nodes and edges, respectively. We also use a nonlinearity \(\sigma_A\) as \citet{velickovic2018graph}, which in our case is the LeakyReLU activation function. The \(\cdot^T\) operator represents transposition, while \(||\) is the concatenation operator.

We obtain the final nodes and edges representations by aggregating the information from the adjacent nodes for each node in \(\boldsymbol{h^{'}_N}\) and the information from the adjacent edges for each node into \(\boldsymbol{h^{'}_E}\):

\begin{equation}
    \label{eq:gatedge}
    \boldsymbol{h^{'}_N} = \text{softmax}(\boldsymbol{e_N}) \boldsymbol{W_Nh_N}
\end{equation}

\begin{equation}
    \label{eq:gatnode}
    \boldsymbol{h^{'}_E} = \text{softmax}(\boldsymbol{e_E})\boldsymbol{W_Eh_E}
\end{equation}

In the end, we combine this information into a single representation for each node into \(\boldsymbol{h^{'}}\) as follows:

\begin{equation}
    \label{eq:gatfin}
    \boldsymbol{h^{'}} = \boldsymbol{h^{'}_N} + \boldsymbol{h^{'}_E} 
\end{equation}

\noindent\textbf{Final Score.} After encoding the graphs, we select the relevant information from the provided knowledge, given the encoded claims. For this, we compute a relevance matrix by calculating the alignment between each encoded claim \(\boldsymbol{h^i_c}\) and all the encoded relations \(\boldsymbol{h^j}\) from the sampled KG (i.e., sub-KG). We calculate the alignment using the cosine similarity:

\begin{equation}
    \label{eq:kgextract}
    \boldsymbol{R^{ij}} = \cos (\boldsymbol{h^i_c}, \boldsymbol{h^j})
\end{equation}
We then use this matrix to finally aggregate all the relevant information from the sub-KG into a matrix containing as many vectors as nodes in the original claim graph, each vector being a numeric encoding representation for every encoded claim node, which in turn represents an embedding for all the neighboring claims:

\begin{equation}
    \label{eq:kgagg}
    \boldsymbol{\Bar{H}} = \boldsymbol{Rh}
\end{equation}
We separately encode the (question, choice) pair into \(\Bar{c}\) and decide what information is better suited for the final decision for the current choice. We employ a self-attention mechanism for this task and provide a score based on the gathered information and the given choice:

\begin{equation}
    \label{eq:selfatt}
    [\boldsymbol{c_{\text{final}}} || \boldsymbol{H_{\text{final}}}] = \text{SelfAttention}([\boldsymbol{\Bar{c}} || \boldsymbol{\Bar{H}}])
\end{equation}

Finally, we compute the score:

\begin{equation}
    \label{eq:graffinal}
    \text{score} = \sigma(\boldsymbol{W_{\text{final}}}\boldsymbol{c_{\text{final}}})
\end{equation}
where \(\sigma\) is the sigmoid logistic activation used to provide a probability score, and \(W_{final}\) is a learnable parameter.

\begin{table*}[t]
  \centering
  \resizebox{\linewidth}{!}{
  \begin{tabular}{lcccccccccc}
    \toprule
    \textbf{Model} & \textbf{Civil} & \textbf{Penal} & \textbf{Civil Proc.} & \textbf{Penal Proc.} &  \textbf{Administrative} & \textbf{Work} & \textbf{Family} & \textbf{International} & \textbf{Commercial} & \textbf{Average} \\
    \midrule
    QBERT & 35.48 & 38.29 & 35.82 & 40.10 & 36.63 & 40.40 & 38.24 & 39.29 & 39.39 & 38.18 \\
    CrossQBERT & 41.94 & 36.04 & 37.31 & 41.15 & 41.58 & 35.35 & 34.31 & 42.86 & 36.36 & 38.54 \\
    ColBERT & 44.09 & 38.29 & 47.76 & 37.50 & 41.58 & 42.42 & 41.18 & 48.81 & 40.40 & 42.45 \\
    LLM & 48.94 & 40.18 & 41.00 & 42.27 & 46.53 & 48.48 & 49.02 & 52.38 & 39.39 & 45.35  \\
    LLM RAG & \textbf{53.19} & 43.75 & 38.81 & 42.78 & 57.43 & 56.57 & 63.73 & 52.38 & \textbf{57.58} & 51.81 \\
    LLM LFT & 45.74 & 51.79 & 74.63 & 48.97 & 52.48 & 49.49 & 54.90 & \textbf{70.24} & 53.53 & 55.75 \\
    \hdashline\noalign{\vskip 0.7ex}
    GRAF(Ours) & 49.46 & \textbf{52.70} & \textbf{78.46} & \textbf{51.05} & \textbf{59.18} & \textbf{57.29} & \textbf{68.69} & 67.12 & 56.84 & \textbf{60.09} \\
    \bottomrule
  \end{tabular}
  }
  \caption{Accuracy results on promotion exams.}
  \label{tab:prom}
\end{table*}

\section{Experiments and Results}

In this section, we present the results of our extensive experimentation and discuss the findings obtained.

\subsection{Baselines}
\label{sec:bibtex}

For encoder models, we adopt approaches similar to those in information retrieval (IR) and retrieval augmented generation (RAG) \citep{wang2024utilizing}. In Appendix \ref{appendix:hyperparameters}, more experimental details are discussed. As baselines, we employ QBERT, ColQBERT \citep{manotumruksa2020crossbert}, ColBERT \citep{khattab2020colbert}, Large Language Models (LLMs) such as FLAN-T5 \cite{raffel2020exploring}, Mistral \cite{jiang2023mistral} and Llama 3.1 8B \cite{dubey2024llama}, LLM with RAG \citep{lewis2020retrieval}, and LLM fine-tuned with Low-Rank Adaptation method (LoRA) \citep{hu2021lora}. More details regarding these baselines can be found in Appendix \ref{appendix:baselines}. For information on the language models employed, see Appendix \ref{appendix:lm}.

\subsection{Evaluation Metric}

We evaluated the models using the score that a model would receive on the given test, equivalent to the model's accuracy on the task. No extra points are given or deducted if the model mispredicts correct answers or fails to include all correct answers. We use this metric to emphasize the actual test performances of the models. Moreover, we argue that the dataset is balanced and suitable for comparative analysis, and thus, we consider this metric sufficient to avoid overwhelming the results section with excessive numbers.


\subsection{Analysis}

\begin{figure}[ht]
  \centering
  \includegraphics[align=c, width=\columnwidth]{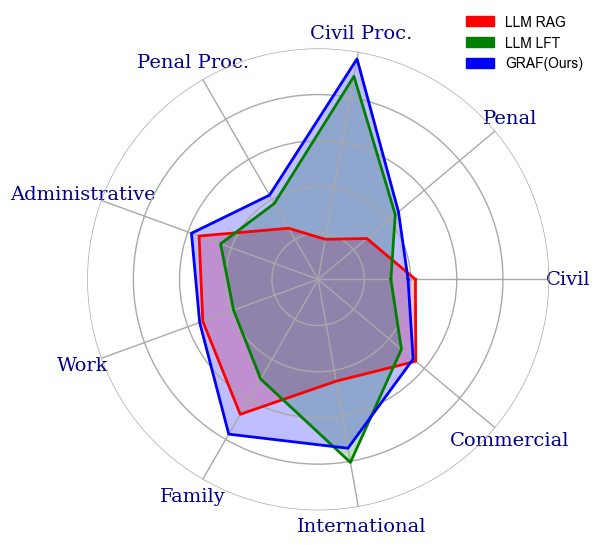}
  \caption{Comparative performances of best baselines and our approach on different law branches.}
  \label{fig:juradar}
\end{figure}

\begin{figure}[t]
  \includegraphics[width=\columnwidth]{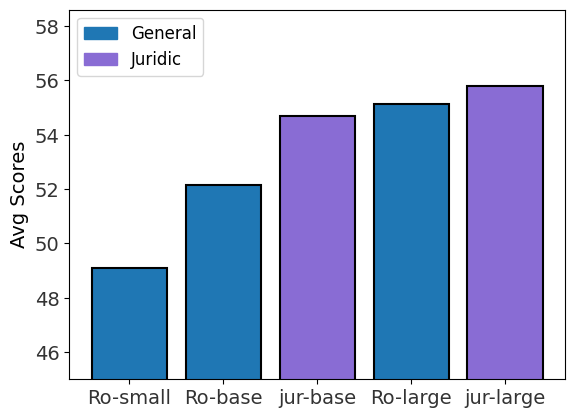}
  \caption{Comparative performances of different backbone encoder models (i.e., BERT-based models) of different sizes and pre-trained on different domains.}
  \label{fig:scoreslaws}
\end{figure}

Our analysis evaluates the performance of our proposed model through quantitative and qualitative evaluations against baseline approaches.
\\ \\
\noindent From a quantitative \textbf{ perspective}, we systematically compare the performance of the model in different legal examinations. All models are evaluated directly for the promotion exams with a single correct answer, as shown in Table \ref{tab:prom}. However, encoder-based models and LLMs are evaluated separately for exams with multiple correct answers to ensure a fair comparison (\S \ref{sec:problem}). Our model outperforms baselines in 6 out of 9 legal branches, despite slight inconsistencies due to training on the entire dataset. Tables \ref{tab:results_entr} and \ref{tab:results_bar} present detailed performance breakdowns for encoder-based and LLM-based models across different examination types, with additional granular results available in Appendix \ref{appendix:detailed_evaluation}.
\\ \\
\noindent From a \textbf{qualitative perspective}, we provide a comparative study for the promotion exams to improve our approach over the baselines shown in Figure \ref{fig:juradar}. We analyze the improvements introduced by our model, particularly in terms of backbone model scaling and domain-specific fine-tuning. As illustrated in Figure \ref{fig:scoreslaws}, pre-trained models for the legal domain significantly outperform their general-purpose counterparts. Our model surpasses baseline encoder models in all evaluation settings, while evidence suggests that poorer performance can be solved by increasing the size of the backbone model. Our framework learns to extract relevant information while answering questions by combining retrieval, fine-tuning, KGs, and inter-entity relationships within legal texts. This structured approach improves performance, allowing fine-tuning while maintaining competitive results in areas where RAG excels. Conversely, though effective in specific legal branches, fine-tuned LLMs struggle to generalize across domains and are susceptible to hallucinations when provided with external context.

We also measure the agreement among LLMs on various categories and topics. We assess the average pairwise percentage agreement (APPA) by computing the percentage of samples in every pair of LLMs responses that produced the same result and then averaging the scores. Table \ref{tab:summary_ppa_and_freiss} presents the APPA for every exam type and category. The values range between 40\% and 50\%, meaning low agreement. However, the Fleiss' $\kappa$ is slightly negative, resulting in no agreement. Therefore, there are questions on which LLMs perform poorly.

Additionally, we provide an in-depth analysis of LLM performance based on question difficulty in Appendix \ref{appendix:data_difficulty}.

\begin{table}[t!]
  \centering
  \resizebox{\columnwidth}{!}{
  \begin{tabular}{lccccc}
    \toprule
    \textbf{Model} & 
    \textbf{Civil} & \textbf{Penal} & \textbf{Civil Proc.} & \textbf{Penal Proc.} &  \textbf{Average} \\
    \midrule
    QBERT & 41.18 & 49.10 & 37.25 & 39.22 & 41.69 \\
    CrossQBERT & 41.18 & 49.02 & 43.14 & 37.25 & 42.65 \\
    ColBERT & 45.10 & 43.14 & 50.98 & 41.18 & 45.10 \\
    LLM & 39.21 & 31.37 & 15.69 & 27.45 & 28.43 \\
    LLM RAG & 43.14 & 27.45 & 21.57 & 27.45 & 29.90 \\
    LLM LFT & 37.25 & 45.10 & 41.18 & 41.18 & 41.18 \\
    \hdashline\noalign{\vskip 0.7ex}
    GRAF(Ours) & \textbf{60.78} & \textbf{62.75} & \textbf{54.90} & \textbf{56.86} & \textbf{58.82} \\
    \bottomrule
  \end{tabular}
  }
  \caption{Accuracy results on entrance exams.}
  \label{tab:results_entr}
\end{table}

\begin{table}[t!]
  \centering
  \resizebox{\columnwidth}{!}{
  \begin{tabular}{lccccc}
    \toprule
    \textbf{Model} & 
    \textbf{Civil} & \textbf{Penal} & \textbf{Civil Proc.} & \textbf{Penal Proc.} &  \textbf{Average} \\
    \midrule
    QBERT & 40.32 & 32.26 & 40.32 & 36.29 & 37.30 \\
    CrossQBERT & 41.94 & 38.52 & 39.52 & 34.68 & 38.67 \\
    ColBERT & 38.71 & 33.87 & 36.29 & 39.52 & 37.10 \\
    LLM & 21.77 & 25.00 & 23.39 & 18.54 & 22.18 \\
    LLM RAG & 26.61 & 29.03 & 33.87 & 20.97 & 27.62 \\
    LLM LFT & 31.45 & 32.26 & 33.06 & 36.29 & 33.27 \\
    \hdashline\noalign{\vskip 0.7ex}
    GRAF(Ours) & \textbf{51.61} & \textbf{61.29} & \textbf{56.10} & \textbf{59.02} & \textbf{57.01} \\
    \bottomrule
  \end{tabular}
  }
  \caption{Accuracy results on bar exams.}
  \label{tab:results_bar}
\end{table}

\begin{table}[ht]
  \centering
  \small
  \begin{tabular}{llcc}
    \toprule
    \textbf{Exam Type} & \textbf{Category} & \textbf{APPA (\%)} & $\kappa$ \\
    \midrule
    \multirow{4}{*}{Entrance} 
    & Civil & 47.65 & -0.0200 \\
    & Penal & 44.51 & -0.0200 \\
    & Civil Procedure & 47.25 & -0.0200 \\
    & Penal Procedure & 45.88 & -0.0200 \\
    \midrule
    \multirow{4}{*}{Bar}
    & Civil & 45.00 & -0.0081 \\
    & Penal & 48.55 & -0.0081 \\
    & Civil Procedure & 49.11 & -0.0081 \\
    & Penal Procedure & 47.50 & -0.0081 \\
    \midrule
    \multirow{9}{*}{Promotion}
    & Administrative & 40.99 & -0.0100 \\
    & Civil & 42.34 & -0.0108 \\
    & Commercial & 41.92 & -0.0102 \\
    & International & 43.10 & -0.0120 \\
    & Penal & 40.18 & -0.0045 \\
    & Civil Procedure & 40.25 & -0.0017 \\
    & Penal Procedure & 43.30 & -0.0052 \\
    & Family & 41.35 & -0.0100 \\
    & Work & 41.01 & -0.0102 \\
    \bottomrule
  \end{tabular}
  \caption{Summary of average pairwise percentage agreement (APPA) and Fleiss' $\kappa$ scores for LLM models.}
  \label{tab:summary_ppa_and_freiss}
\end{table}

\begin{table}[t]
  \centering
  \small
  \begin{tabular}{lc}
    \toprule
    \textbf{Model} & \textbf{Avg. Accuracy} \\
    \midrule
    GRAF &  55.61\\
    \quad\textit{w/o claim graph} & 53.14\(_{\downarrow2.47}\)\\
    \quad\textit{w/o KG} & 53.61\(_{\downarrow2.00}\) \\
    \quad\textit{w/o claim graph + emb. sum} & 49.51\(_{\downarrow6.10}\) \\
    \bottomrule
  \end{tabular}
  \caption{Results for different ablated components of our framework on the Promotion exams.}
  \label{tab:ablation}
\end{table}

\subsection{Ablation Study}

In Table \ref{tab:ablation}, we present the effect of removing different parts of our algorithm in an ablation study conducted on the promotion exams. We remove the claim graph and the KG, and we experiment with collapsing the KG embeddings via summation (denoted \textit{emb. sum}). In this sense, we experiment with various ablated components and found that missing KG or claim graphs underperform the LLM baselines. Even when the model uses only the KG, its performance is poorer because irrelevant information may serve as a distractor in the decision-making process. We show that our claim-aware information extraction mechanism is effective in enhancing retrieval capabilities and, consequently, overall performance. Merely collapsing the extracted information via summation would still not compensate for the missing claim-aware IR, but would significantly damage the model's performance.

\section{Conclusions}

In this paper, we have introduced new resources for the Romanian legal domain, being the first to construct and release a Romanian legal KG. We also proposed a novel approach for KG IR, namely GRAF, which surpassed the baselines proposed for MCQA. \\
\\
\noindent\textbf{Future Work.}
We expect solutions that will improve upon our proposed method in every possible aspect. Additionally, there may be solutions that could potentially explore dataset augmentation using LLMs. Studies could be conducted on target domain IR, which may include multiple languages, and JuRO, CROL, and even the KG Law-RoG could represent good foundational resources. We hope that our work will motivate further exploration of underrepresented languages and, in turn, inspire the development of solutions that work in low-resource settings.

\section*{Limitations}

We have released a legal MCQA dataset by gathering questions from all available law examinations nationwide, providing sufficient samples for training. However, it may not be enough for training in a single law branch, which is why we opted for training on the entire dataset.

The goal of our work is to enhance resources and develop a methodological approach to answering legal questions. Since such systems are meant to help users understand the law, they are not yet entirely accurate. The best average score achieved by our approach is only 60\%. Therefore, further research is required in this direction, as the legal domain is a sensitive topic when considering the application of machine learning systems in the assessment of laws. We believe that deploying such systems would require human validation by legal experts to minimize the risk of providing unlawful responses.

\section*{Ethical Considerations}

We have collected our dataset from various official public portals. To protect this dataset from improper use, we have decided to license its use solely for research purposes. It should not be used in commercial settings under any circumstances. Our work was performed in a manner that did not rely on external human crowd-workers and did not raise any ethical concerns. The data do not contain sensitive personal information that could identify any real person.
Anonymized abbreviations are used in all of the hypothetical presented scenarios rather than any person's name. Since the data was collected from the public domain and made available by applicable law by the administrative institutions in question, we release these resources under the CC BY-NC-SA 4.0 license\footnote{\url{https://creativecommons.org/licenses/by-nc-sa/4.0/}}, allowed by the current European regulations\footnote{\url{https://eur-lex.europa.eu/eli/dir/2019/790/oj}}.

\section*{Acknowledgements}
This work was supported by the National University of Science and Technology POLITEHNICA Bucharest through the PubArt program.

\bibliography{custom}

\clearpage

\appendix

\section{Dataset Analysis}\label{appendix:analysis}

\begin{figure*}[!htp]
  \includegraphics[align=c, width=0.48\linewidth]{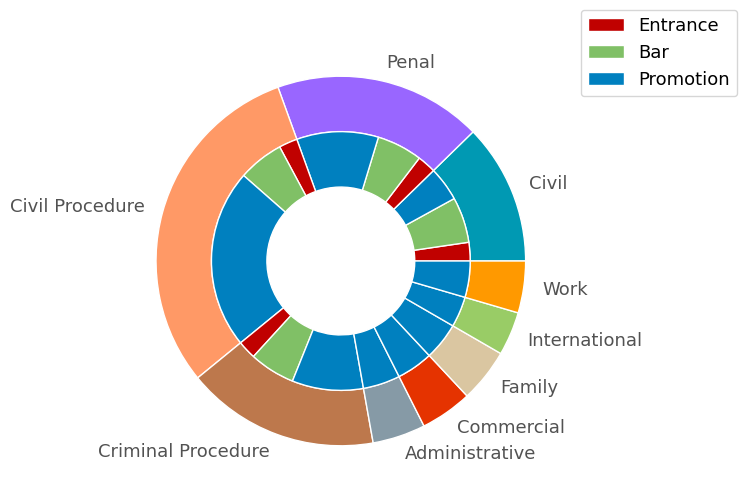} \hfill
  \includegraphics[align=c, width=0.48\linewidth]{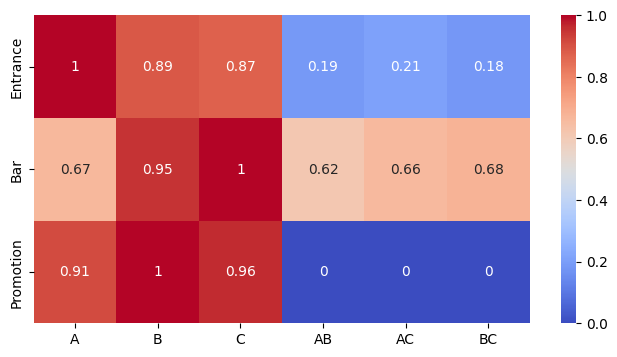}
  \caption{On the left, the number of samples from each law category in the JuRO dataset. On the right, the class equilibrium is depicted via color variations in the heatmap. The heatmap scores are normalized by dividing each value by the maximum of the respective row.}
  \label{fig:juropie}
\end{figure*}

The domain distribution of the JuRO, along with the distribution of the answers, is presented in Figure~\ref{fig:juropie}. Because of the examinations' format, at most two answers are correct. However, in the case of promotion exams, only one answer is correct. The domains of the questions are \textit{civil procedure}, \textit{penal procedure}, \textit{penal}, \textit{civil}, \textit{work}, \textit{administration}, \textit{commercial}, \textit{family}, and \textit{international}.

Table \ref{tab:tfjuro} presents the TF-IDF scores~\cite{salton1975vector} for JuRO dataset, calculated using the following formula:

\begin{equation}
    \resizebox{0.8\columnwidth}{!}{
    $\text{score}(t, C)=\frac{f(t,C)}{|\{w | d \in C, w \in d\}|} \log\frac{|C|}{|\{d | d \in C\ and\ t \in d\}|}$
    \label{eq:score}
    }
\end{equation}
where:
\begin{itemize}
	\item the current term for which we compute the score is denoted by \(t\);
    \item \(C\) is the corpus of documents, each document containing multiple words;
    \item \(f(t,C)\) is the frequency of the term \(t\) relative to the corpus \(C\).
\end{itemize}

We notice a high score for the word ``penal'' compared to other words, indicating a possible prevalence of penal-related content in the dataset. Moreover, the terms such as ``case'', ``term'', ``appeal'', ``judgement'', ``court'', and ``request'' indicate procedures.

\begin{table}[!ht]
    \centering
    \small
    \begin{tabular}{llc}
        \toprule
        \textbf{Word} & \textbf{Translation} & \textbf{TF-IDF Score}\\
        \midrule
         cerere & request & 0.03507 \\
         instanţă & court & 0.03352 \\
         caz & case & 0.02769 \\
         lege & law & 0.02627 \\
         penal & penal   & 0.02622 \\
         hotărâre & decision & 0.02614 \\
         termen & term & 0.02602 \\
         apel & appeal & 0.02565 \\
         sine & self & 0.02553 \\
         judecată & judgement & 0.02486 \\
         \bottomrule
         
      \end{tabular}
\caption{TF-IDF scores for the top ten words in the JuRO dataset.}
\label{tab:tfjuro}
\end{table}

In Table \ref{tab:tfcrol}, we report the TF-IDF scores for the CROL corpus. Generally, there are words commonly found in articles, but no word indicates a significant bias towards some specific legal area.

\begin{table}[ht]
  \centering
    \small
    \begin{tabular}{llc}
        \toprule
        \textbf{Word} & \textbf{Translation} & \textbf{TF-IDF Score}\\
        \midrule
         articol & article & 0.03262 \\
         caz & case & 0.02550 \\
         persoană & person & 0.02510 \\
         lege & law & 0.02471 \\
         bun & good & 0.02429 \\
         alineat & paragraph & 0.02314 \\
         prevedea & stipulate & 0.02260 \\
         drept & just/correct/law & 0.02213 \\
         număr & number & 0.01965 \\
         public & public & 0.01961 \\
         \bottomrule
    \end{tabular}
    \caption{TF-IDF Scores for the top ten words from the CROL corpus.}
    \label{tab:tfcrol}
\end{table}

\begin{figure*}[ht!]
  \includegraphics[width=\linewidth]{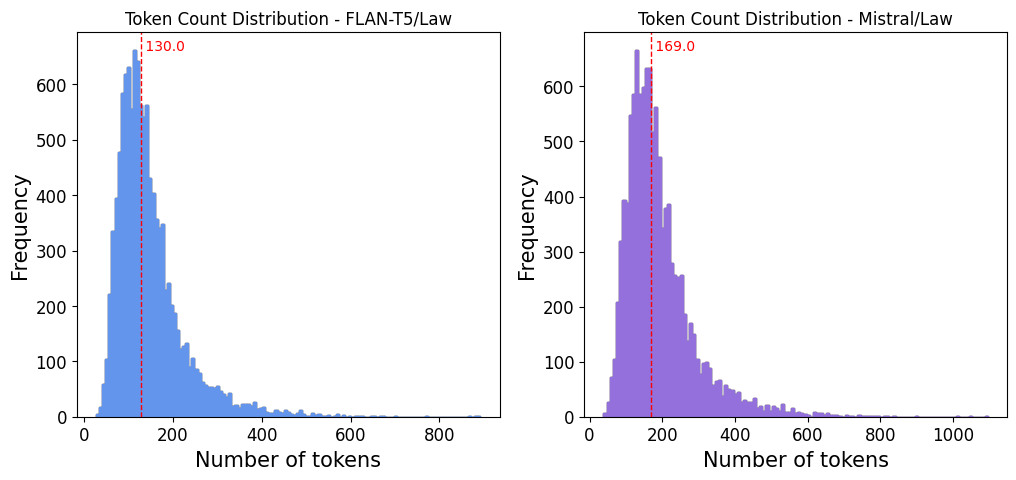}
  \caption {The token distribution on the JuRO dataset for Flan-T5 and Mistral tokenizers.}
  \label{fig:disttoken}
\end{figure*}

In Figure \ref{fig:disttoken}, we show the token distribution for both the FLAN-T5 \citep{raffel2020exploring} and Mistral \citep{jiang2023mistral} models using their tokenizers. The distributions behave approximately the same; the difference is that the Mistral tokenizer tends to use more tokens to represent the text than the FLAN-T5 one. Table \ref{tab:jurocount} shows a detailed distribution of questions for each examination. 

\section{Topic Analysis}
\label{appendix:topic_analysis}

\begin{figure*}[!ht]
  \centering
  \includegraphics[trim={4cm 2cm 1cm 2cm},clip,width=\textwidth]{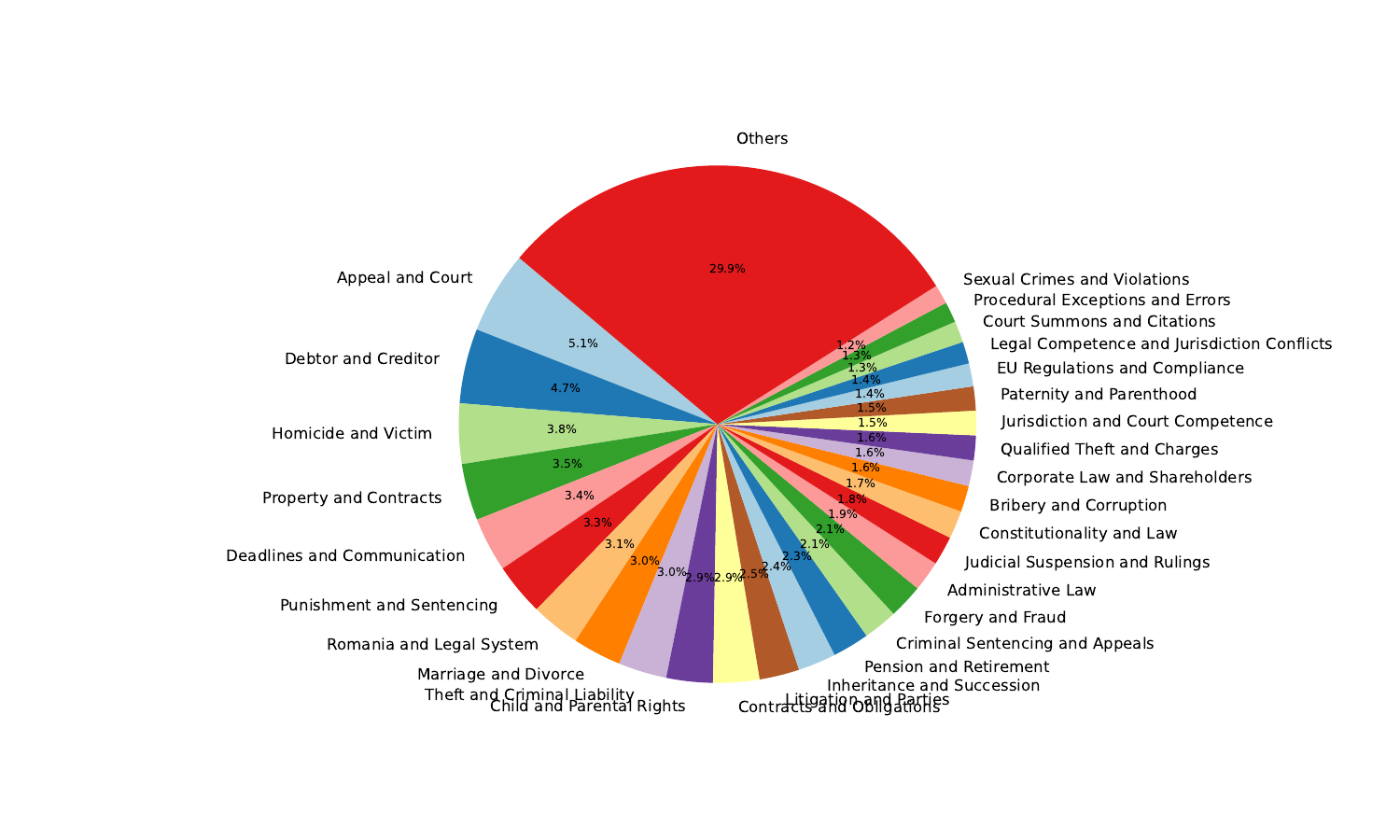}
  \caption{The distribution of top 30 topics in the JuRO dataset.}
  \label{fig:topics_distrib_juro}
\end{figure*}

\begin{figure*}[!ht]
  \centering
  \includegraphics[trim={2.5cm 2cm 0.5cm 2cm},clip,width=\textwidth]{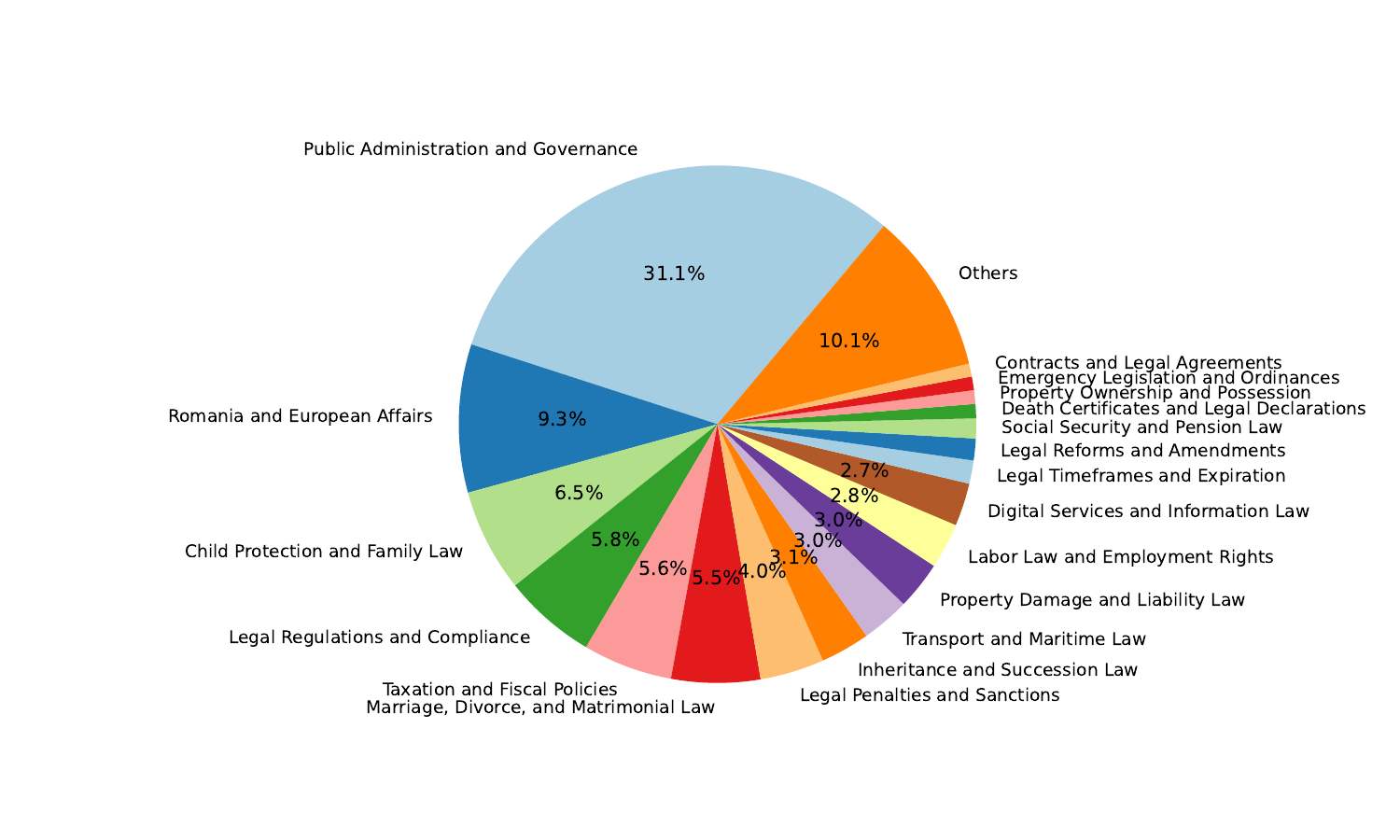}
  \caption{The distribution of top 20 topics in the CROL dataset.}
  \label{fig:topics_distrib_crol}
\end{figure*}

To better understand the performance of the LLMs used in our work, we extracted the main topics from the CROL and JuRO datasets and present the performance relative to these. For both datasets, the topic extraction procedure is similar. First, we preprocess the JuRO dataset by merging each question with the set of answer choices. In the case of CROL, we perform a minimal data cleaning procedure to remove frequent words and structures that do not represent topics such as law numbers (e.g., Arabic numerals, Roman numerals, references to paragraphs or other laws like ``lit. (a)'', months of the year, and dates), separators, and the repealed laws, since they have a very similar formulation and any line shorter than five characters. Then, we employ BERTopic~\cite{grootendorst2022bertopic}, which generates transformer-based embeddings and class-based TF-IDF to create dense clusters of semantically similar documents. We set the language to Romanian to output 100 topics and a fixed random seed for reproducibility. The other parameters were left to their default values. We remove outliers and topics that contain only stopwords from the resulting output.

The extracted topics for the JuRO dataset are presented in Table \ref{tab:top_30_juro_ro} in the original Romanian language and Table \ref{tab:top_30_juro_en} for the topics and keywords translated into English. Similarly, Tables \ref{tab:top_30_crol_ro} and \ref{tab:top_30_crol_en} present the top 30 topics of the CROL corpus in Romanian and English, respectively. Both datasets cover a wide variety of topics in the legal domain, ranging from \textit{Appeal and Court}, \textit{Punishment and Sentencing}, \textit{Romanian Legal System}, to \textit{Public Administration and Governance}, \textit{Child Protection and Family Law}, \textit{Labor Law and Employment Rights}, and many other legal subjects. We provide the distribution of those main topics in Figures \ref{fig:topics_distrib_juro} and \ref{fig:topics_distrib_crol} for JuRO and CROL, respectively. Most of the topics represent a small percentage of the datasets, emphasizing the large diversity of topics addressed in our proposed resources.

\section{Dataset Difficulty}
\label{appendix:data_difficulty}

Inspired by other works \cite{zhengllmjudge2023,muennighoff2025s1}, we estimate the question difficulty from the JuRO dataset by analyzing the LLMs' performance (i.e., using the LLMs to judge the difficulty of the questions). 

\begin{figure*}[t]
  \centering
  \includegraphics[width=\textwidth]{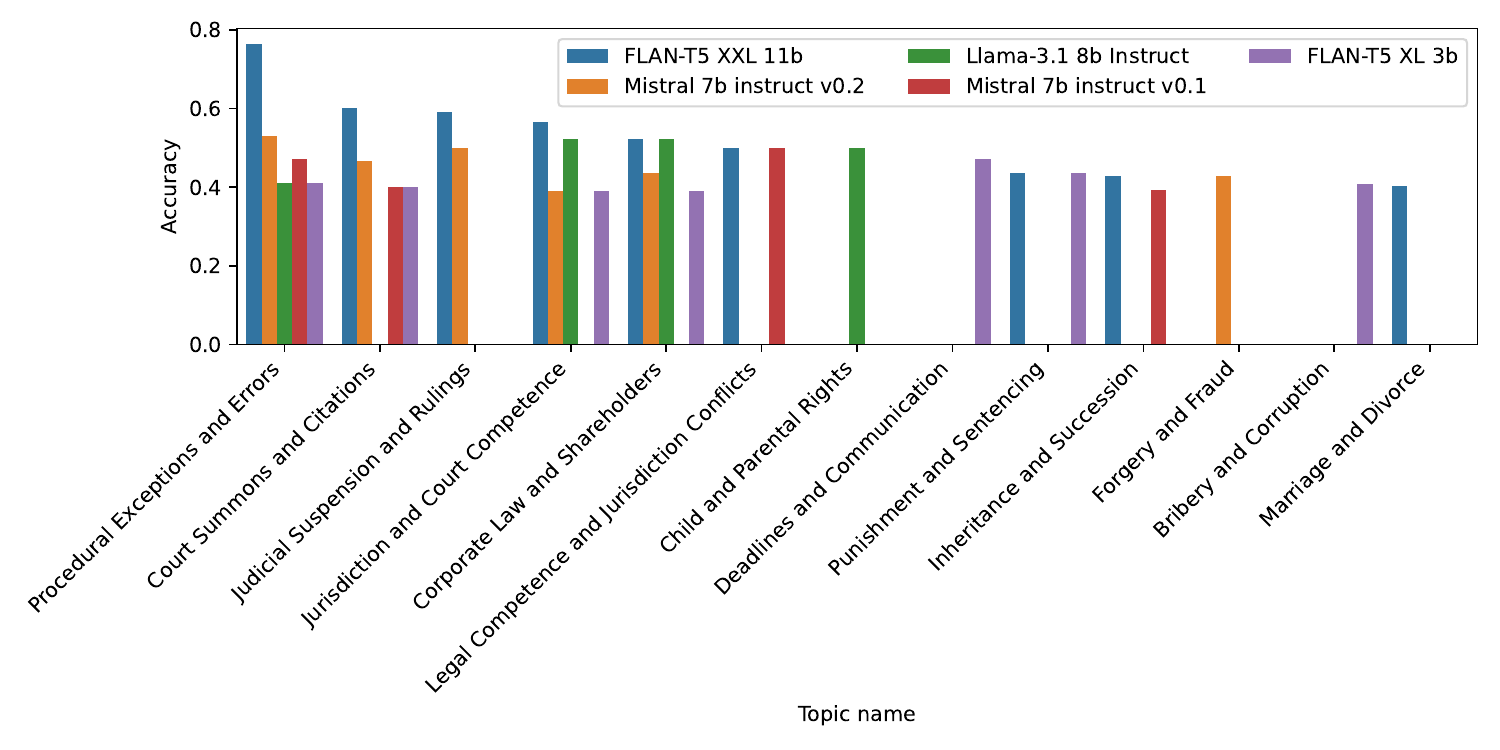}
  \caption{Accuracy computed for samples in the top 13 topics from the JuRO dataset, for every LLM. Higher is better.}
  \label{fig:accuracy_per_topic_model}
\end{figure*}

We base our approach on the topics identified in Appendix \ref{appendix:topic_analysis}. Breaking down the model-level results in Figure~\ref{fig:accuracy_per_topic_model}, we notice that FLAN-T5 XXL performs the best on \textit{procedural exceptions}, outperforming other models by 20-40\%. However, there are topics where some models did not answer any question correctly, such as \textit{marriage and divorce}, \textit{bribery and corruption}, and \textit{fraud}. 

\begin{figure*}[t]
  \centering
  \includegraphics[width=\textwidth]{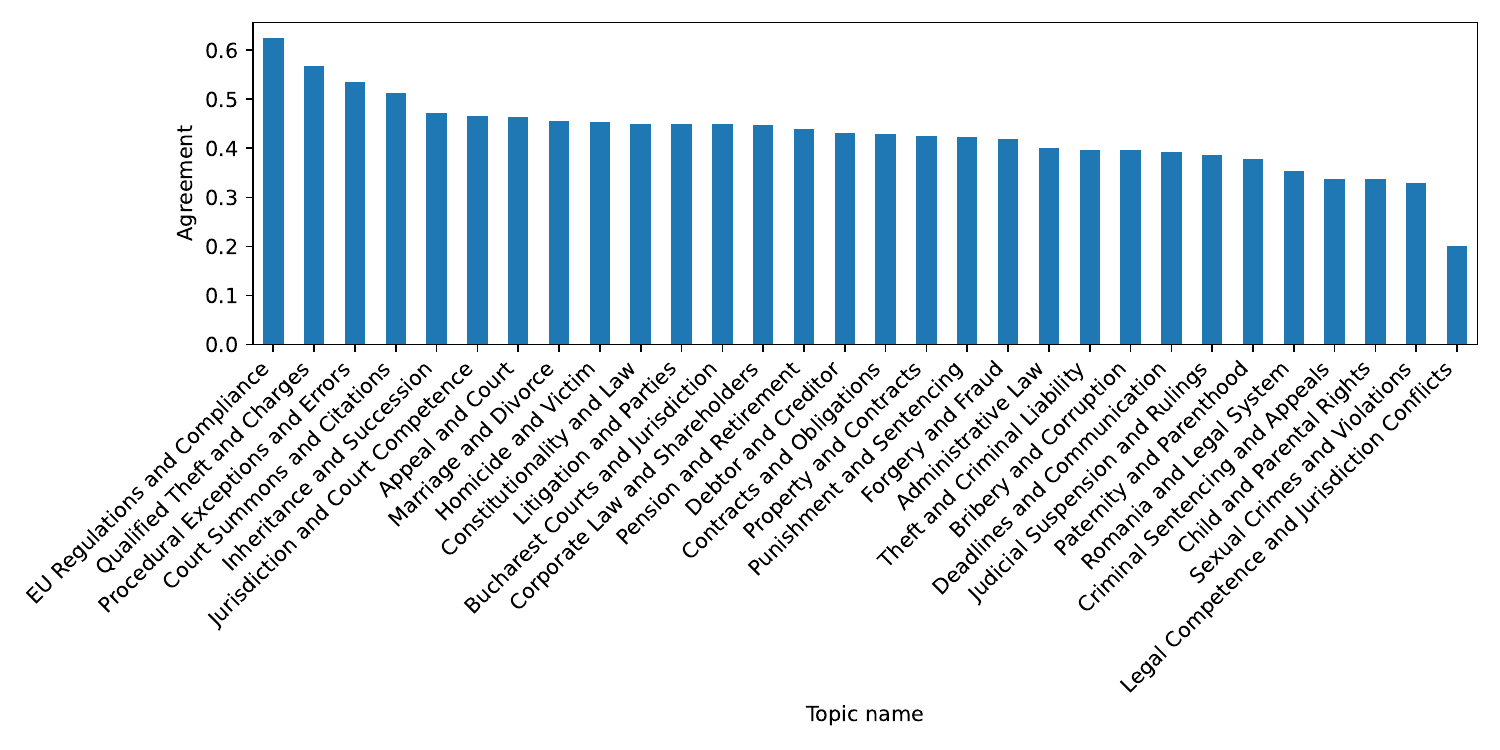}
  \caption{Per topic average pairwise percentage agreement scores when employing Llama-3.1 8B Instruct, FLAN-T5 XL, FLAN-T5 XXL, Mistral 7B Instruct v0.1, and Mistral 7B Instruct v0.2 LLMs. Higher is better.}
  \label{fig:agreement_per_topic}
\end{figure*}

We also decompose the APPA score for every topic in Figure \ref{fig:agreement_per_topic} to identify situations where the models perform poorly. We observe that the models yield better results on questions related to \textit{EU Regulations} and \textit{theft}, while performing poorly on subjects such as \textit{sexual crimes} and \textit{jurisdiction conflicts}. However, the agreement is below 50\% for most topics.

Additionally, we estimate the difficulty of each question per topic based on the model's performance. We normalize performance per model to account for the fact that some models perform better than others. If a better model fails on a question that weaker models also fail on, the question is likely to be more difficult. Formally, for every $i$ sample, we first compute the performance score $score_{i,m}$ assigning 1 for every correct prediction with ground truth and 0 otherwise for every model experiment $m$. Then we calculate the overall per-model performance $\mu_m = \mean_{i} (score_{i,m})$ and standard deviation $\sigma_m = \std_{i} (score_{i,m})$ for every model $m \in M$ across all topics. For every prediction $i$ associated with a model $m$, the z-score is defined as:

\begin{equation}
\mathop{\mbox{$z$-$\mathit{score}_{i,m}$}} = \frac{score_{i,m} - \mu_m}{\sigma_m}
\end{equation}

Then, to compute the topic-based z-score, we average the z-scores within a given topic $t$ for all models $m \in M$:

\begin{equation}
\mathop{\mbox{$z$-$\mathit{score}_{t}$}} = \frac{1}{|t| \cdot |M|} \sum_{i \in t} \sum_{m \in M}\mathop{\mbox{$z$-$\mathit{score}_{i,m}$}}
\end{equation}

\begin{figure*}[t]
  \centering
  \includegraphics[width=\textwidth]{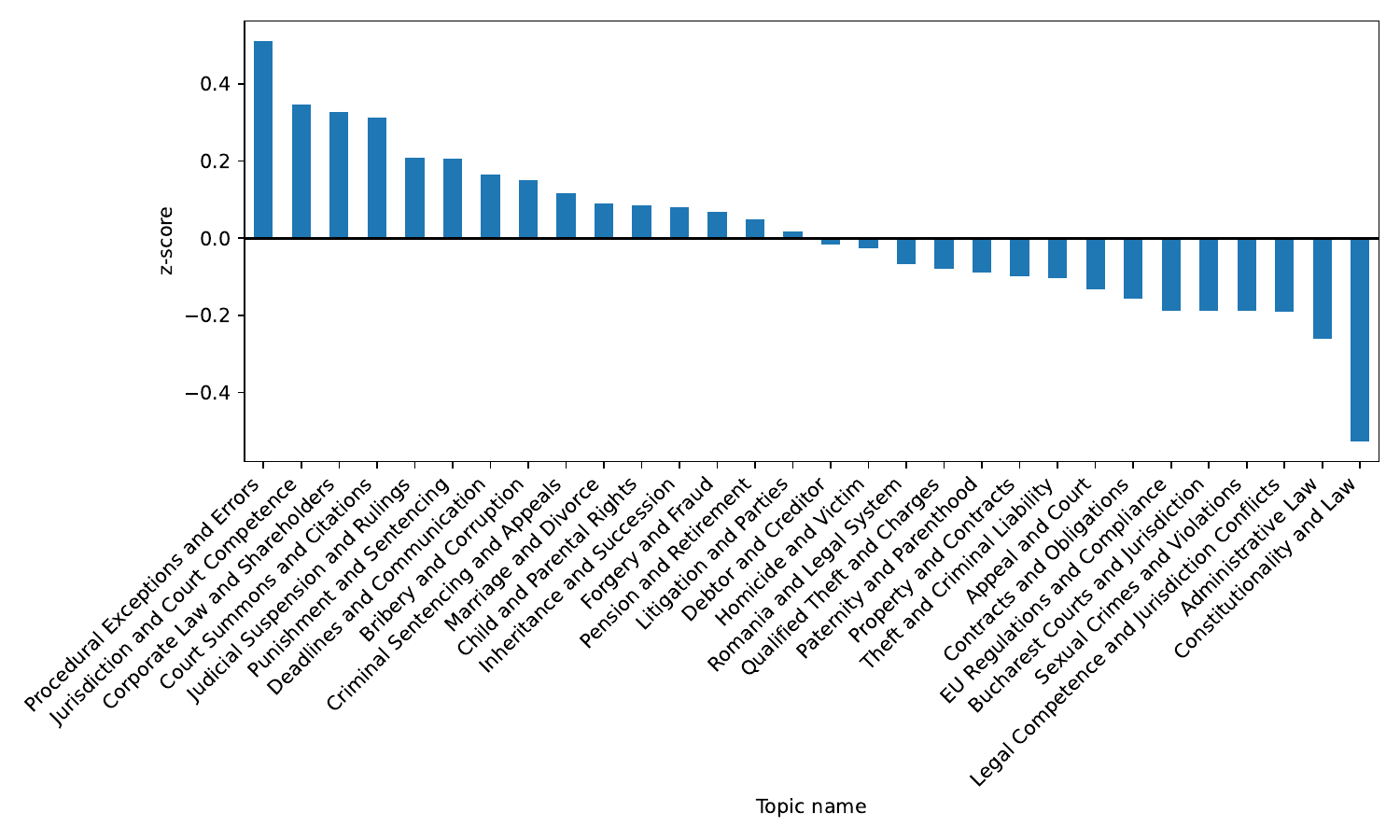}
  \caption{Per-topic question difficulty on the JuRO dataset relative to LLM performance using the z-score normalization. High positive values indicate that the questions from the given topic are easier, while lower negative values indicate that the questions from a given topic are more difficult.}
  \label{fig:difficulty_scores}
\end{figure*}

The final z-scores are shown for the most frequent topics in Figure~\ref{fig:difficulty_scores}. The most straightforward topics from the LLM perspective are \textit{procedural exceptions and errors}, \textit{jurisdiction and court competence}, \textit{corporate law}, and \textit{court summons and citations}. On the other end of the spectrum, the most challenging questions were related to \textit{constitutional} and \textit{administrative} laws.

\begin{figure*}[t]
  \centering
  \includegraphics[width=\textwidth]{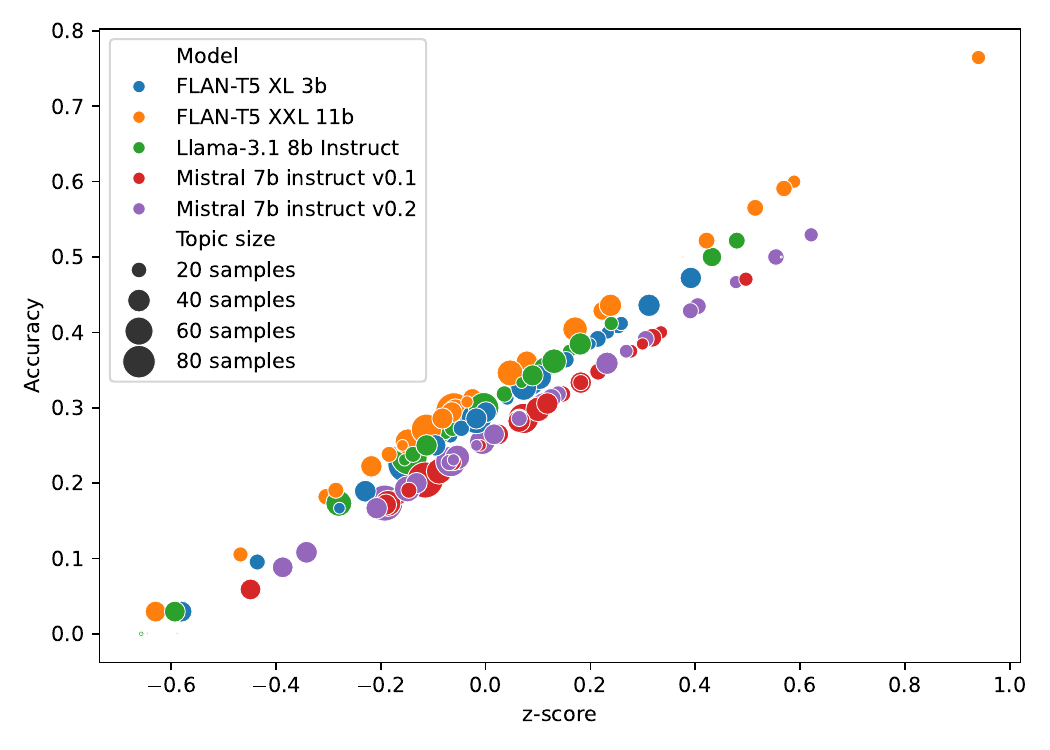}
  \caption{The dependency between accuracy, question difficulty (as z-score), model, and topic size. Larger language models having the Romanian language in the training set (i.e., FLAN-T5) perform better than smaller models trained on English-only data (i.e., Mistral 7B). Most topics reside in the medium to higher difficulty levels from the LLM performance perspective (i.e., z-score less than 0), achieving lower accuracy on those topics (i.e., under 40\%). There is a single exception for FLAN-T5 XXL on \textit{Procedural Exceptions and Errors}, achieving 76\% with a z-score of 0.94. At the bottom of the scale, the models perform worse at \textit{Constitutionality and Law} and \textit{Legal Competence and Jurisdiction Conflicts} topics.}
  \label{fig:zscore_vs_accuracy}
\end{figure*}

We present a multi-dimensional analysis in Figure \ref{fig:zscore_vs_accuracy} considering model accuracy, z-score-based question difficulty, and topic size. We demonstrate that model performance is influenced by the pre-training dataset (i.e., whether it includes the Romanian language), the number of parameters, and the difficulty of the questions. 

\section{Model Architectures}
\label{appendix:arch}

\quad\textbf{Autoregressive Models.} These models are known to exhibit impressive capabilities in generative tasks. They can also be adapted to classification tasks by teaching them the correlation between the class concept and the chosen class symbol. Their goal is to minimize the negative log-likelihood of the class symbol given the input. Specifically:

\begin{equation}
  \label{eq:nll}
  \mathcal{L} = -\sum \log P(t^j_i | \mathcal{Z}(Q_i, C_i, t^k_i); \boldsymbol{\theta})
\end{equation}
where \(j > k\) and \(t^0_i\) represents the empty sequence. The \(\mathcal{Z}\) function maps a given triplet to a sequence that can be processed by the given probabilistic model, known as the model prompt. The prompt serves to facilitate and guide the model towards a lower on-average negative log-likelihood, and consequently, the correct answer. Although our work also explores sequence-to-sequence models, the ones we chose in particular generate output in an autoregressive manner via the decoder module; thus, our previous discussion still holds.

\textbf{Encoder Models.} These models showed excellent performance on classification tasks despite their relatively smaller size in practice. They feature a good semantic understanding of a given sequence via their pre-training objectives. For instance, BERT \citep{devlin2018bert} featured word- and sentence-level pre-training, which allowed it to gain a semantic understanding of language. However, they do not exhibit symbol-level correlation \citep{robinson2023leveraging}, unlike LLMs, and thus, we resort to using their semantic understanding of textual sequences to output a number that represents the degree to which a given choice is correct given a question. We consider two learning goals for these models, the binary cross-entropy minimization for models outputting probabilities:

\begin{equation}
  \label{eq:ce}
  \mathcal{L}_1 = -(o^j_i\log(y^j_i) + (1 - o^j_i)\log(1 - y^j_i))
\end{equation}
where \(o^j_i = \mathbbm{1}_{T_i}(C^j_i)\) is the ground truth and \(y^j_i\) is the model output probability. We also use the cosine similarity embedding loss to align a given question with the correct answer choice:

\begin{equation}
    \label{eq:cs}
    \mathcal{L}_2 = (1 + o^j_i) (1 - y^j_i) + (1 - o^j_i)y^j_i
\end{equation}
where \(o^j_i\) has the same meaning as above except the negative class becomes -1 instead of 0, whereas \(y^j_i = \cos(\Bar{Q_i}, \Bar{C^j_i})\) and we refer to the bar notation as the embeddings of the question and choice respectively.

During inference, we consider the question along with the set of choices and select the top \(|T_i|\) scores in the following way:

\begin{equation}
    \label{eq:top}
    Y_i^* = \text{TopK}(Y_i, |T_i|)
\end{equation}
where \(Y_i = \{y^j_i | y^j_i = \text{Score}(Q_i, C^j_i)\}\). TopK is a generalized argmax function that selects the best \(K\) candidates from a given list. In the end, the chosen options by the model are \(C^k_i\) with \(k \in Y_i^*\).

\section{Experimental Setup}
\label{appendix:hyperparameters}

Training was performed on the entire JuRO dataset for each model and, for testing, we considered the checkpoint with the best evaluation results obtained during the training phase. For encoders, we used BERT-based models that were trained for 50 epochs, even though in almost all cases, the best model was found around epoch 10. We used a learning rate of \(10^{-7}\) and the AdamW \cite{DBLP:journals/corr/KingmaB14} optimizer via vanilla \textit{PyTorch}. All BERT models were fine-tuned on all parameters. LLMs were fine-tuned for 50 epochs using the \textit{Trainer API} provided by the \textit{transformers} library using a learning rate of \(10^{-7}\), AdamW optimizer, LoRA \cite{DBLP:conf/iclr/HuSWALWWC22} alpha of 32, LoRA rank 64, and 2 warm-up steps. All of our experiments were performed on a single NVIDIA A100 80GB to which we had limited and restricted access. We report the results of a single run.

\section{Baselines}
\label{appendix:baselines}

\quad\textbf{QBERT.} We consider the BERT model \citep{devlin2018bert} and construct the input to the LM by appending the given question and the choice in the following way: \textsc{[CLS] + question + [SEP] + choice + [SEP]}. We then use the classification token to attach a fast forward network (FFN) on top with a sigmoid activation function, which will report a score between 0 and 1 for the correctness of the answer choice.

\textbf{CrossQBERT.} As proposed by \citet{manotumruksa2020crossbert}, we proceed by taking the question and the entire set of possible choices and concatenating them in the same fashion as for QBERT. We consider the first three separator tokens and a single FFN, with a sigmoid activation function, which outputs three scores for the same question corresponding to each answer choice. In this way, we provide BERT with more context to gather additional information about neighboring choices, allowing a better and more informed decision.

\textbf{ColBERT.} Initially, an architecture used for information retrieval tasks\citep{khattab2020colbert}, we use it for our task because of its underlying philosophy: aligning textual representations. Thus, we use a model to encode the question and a model to encode the individual choice, and we use the resulting embeddings to perform cosine similarity.

\textbf{LLMs.} We use the generalization capabilities of the LLMs \citep{zhao2023survey}, having decent performances on tasks in no-data settings and no further training. We perform prompt engineering \citep{chen2023unleashing} and extensively experiment with multiple prompts, ultimately providing the results for the prompts that obtain the best performance. For prompts, see Appendix \ref{appendix:ro_prompts} and the translations in Appendix \ref{appendix:en_prompts}.

\textbf{LLM RAG.} We use Retrieval Augmented Generation (RAG) \cite{gao2023retrieval} to provide LLMs with contextual information that would answer the question or guide the model towards the answer. We employ the BM25 retriever \citep{robertson1976relevance} along with the \textit{SpaCy} package for text normalization (tokenization and lemmatization) to extract the top 10 most relevant documents from the corpus. We take the articles from the CROL corpus and split them into 50-word documents. We allow consecutive chunks to overlap by 25 words to maintain context and avoid abrupt disruptions to the flow of information.

\textbf{LLM LFT.} Finally, for our experiments, we fine-tune these LLMs using the LoRA \citep{hu2021lora} adaptation method, which was experimentally shown to match the performance of classic full parameter fine-tuning. This, together with the previous baseline, achieves the best results among the baselines. We opt for the LoRA strategy, since our computational resources would not allow a full       arameter fine-tuning of all our proposed LLMs.

\section{Language Models}
\label{appendix:lm}

The BERT models that we used in our work are the RoBERT \citep{masala-etal-2020-robert}, a Romanian BERT model trained on the general domain, and jurBERT \citep{masala2021jurbert} -- a Romanian BERT model trained on the legal domain. After attempting multiple LLM models and comparing their preliminary performances on this task, we identified the best-performing LLMs, which were of reasonable size for our resources, for the conducted experiments. We used Flan-T5 \citep{raffel2020exploring}, XL 3B and XXL 11B variants, Mistral 7B Instruct \citep{jiang2023mistral}, and Llama 3.1 8B \citep{dubey2024llama}. These LLMs are instruction-tuned, and we opt for this type of LLM for its better performance on instruction-following and target tasks \citep{wei2022finetuned}. We could leverage their initial performance for further fine-tuning. For the GAT model, we used six attention heads, as we experimentally observed that this value represents an equilibrium between the average number of non-zero entries and computational demands when the GAT is initialized and tested with given inputs, aiming to potentially mitigate the dying gradient phenomenon caused by null entries. For KG construction and claim extraction, we used the Mixtral-8x7B-Instruct LLM \citep{jiang2024mixtral} quantized to 8 bits using the int8 algorithm \citep{dettmers20228bit} implemented in the \textit{bitsandbytes} library. Although this model is relatively large, we utilized it in the KG construction and claim extraction process as a trustworthy means, which is more likely to correctly extract entities and relations. More lightweight solutions can be built by training a smaller language model for this task or by distilling \citep{hinton2015distilling, gu2023knowledge} Mixtral to a small language model (SLM), for example. Mixtral did not contribute in any way to helping the rest of our framework make the right choice. It had a very clear and definite role in our algorithm, which can be easily replaced with any other lightweight solution that we could not implement due to the lack of available data for the Romanian language in this sense.

\section{Detailed Evaluation}
\label{appendix:detailed_evaluation}

Tables \ref{tab:allprom}, \ref{tab:allentrance}, \ref{tab:allbar} show all our extensive evaluations conducted on different backbone models and different prompts. P1 and P2 refer to the best prompts for the Mistral and FLAN-T5 models, respectively. Our approach surpasses all the baseline combinations using jurBERT-large \citep{masala2021jurbert} as our backbone encoder model. Moreover, it outperforms encoder models in all settings.

\begin{table*}[t]
  \centering
  \begin{tabular}{llc}
    \toprule
    \textbf{Task} & \textbf{Training/Test/Validation} & \textbf{\# Classes}\\
    \midrule
     \textit{Civil} & 933/266/136 & 3/6 \\
     \quad Entrance & 175/50/26 & 6 \\
     \quad Bar & 432/123/63 & 6 \\
     \quad Promotion & 326/93/47 & 3 \\
     \midrule
     \textit{Penal} & 1380/393/200 & 3/6 \\
     \quad Entrance & 175/50/26 & 6 \\
     \quad Bar & 430/123/63 & 6 \\
     \quad Promotion & 775/220/111 & 3 \\
     \midrule
     \textit{Civil Procedure} & 2070/293/91 & 3/6 \\
     \quad Entrance & 174/50/26 & 6 \\
     \quad Bar & 431/123/63 & 6 \\
     \quad Promotion & 207/67/29 & 3 \\
     \midrule
     \textit{Penal Procedure} & 1282/296/186 & 3/6 \\
     \quad Entrance & 174/50/26 & 6 \\
     \quad Bar & 432/123/63 & 6 \\
     \quad Promotion & 676/123/97 & 3 \\
     \midrule
     \midrule
     \textit{Other Promotion Exams} & 1339/376/195 & 3 \\
     \quad Administrative & 351/99/51 & 3 \\
     \quad Commercial & 344/98/50 & 3 \\
     \quad Family & 355/96/52 & 3 \\
     \quad International & 289/83/42 & 3 \\
     \quad Work & 343/95/50 & 3 \\
     \bottomrule
  \end{tabular}
  \caption{A detailed view of the JuRO dataset regarding the sample distribution among Romanian legal exams and split into training/test/validation sets.}
  \label{tab:jurocount}
\end{table*}

\clearpage

\begin{table*}
  \centering
  \tiny
  \begin{tabular}{lcccccccccc}
    \toprule
    \textbf{Model} & 
    \textbf{Civil} & \textbf{Penal} & \textbf{Civil Proc.} & \textbf{Penal Proc.} &  \textbf{Administrative} & \textbf{Work} & \textbf{Family} & \textbf{International} & \textbf{Commercial} & \textbf{Average}
    \\
    \midrule
    \textit{QBERT} &&&&&&&&&&\\
    RoBERT-small & 29.03 & 38.29 & 34.33 & 29.17 & 30.69 & 31.31 & 31.37 & 27.38 & 39.39 & 32.33 \\
    RoBERT-base & 25.81 & 37.39 & 34.33 & 32.29 & 36.63 & 37.37 & 37.25 & 27.38 & 36.36 & 33.87 \\
    RoBERT-large & 33.33 & 35.14 & 29.85 & 35.94 & 27.72 & 31.31 & 38.24 & 30.95 & 36.36 & 32.20 \\
    jurBERT-base & 35.48 & 28.38 & 35.82 & 37.50 & 25.74 & 40.40 & 29.41 & 28.57 & 30.30 & 32.40 \\
    jurBERT-large & 32.26 & 34.68 & 31.34 & 40.10 & 25.74 & 27.27 & 29.41 & 39.29 & 39.39 & 29.94 \\
    \midrule
    \textit{CrossQBERT} &&&&&&&&&&\\
    RoBERT-small & 35.48 & 33.78 & 32.84 & 38.54 & 41.58 & 26.26 & 28.43 & 42.86 & 35.35 & 35.01 \\
    RoBERT-base & 41.94 & 28.83 & 28.36 & 28.65 & 36.63 & 33.33 & 28.43 & 26.19 & 30.30 & 31.41 \\
    RoBERT-large & 41.94 & 29.28 & 34.33 & 29.69 & 31.68 & 29.29 & 24.51 & 35.71 & 35.35 & 32.42 \\
    jurBERT-base & 41.94 & 31.08 & 37.31 & 35.42 & 30.69 & 35.35 & 29.41 & 30.95 & 36.36 & 34.28 \\
    jurBERT-large & 29.03 & 36.04 & 35.82 & 41.15 & 30.69 & 35.35 & 34.31 & 26.19 & 34.34 & 33.66 \\
    \midrule
    \textit{ColBERT} &&&&&&&&&&\\
    RoBERT-small & 33.33 & 31.98 & 32.83 & 36.46 & 32.67 & 33.33 & 37.25 & 44.05 & 34.34 & 35.14 \\
    RoBERT-base & 40.86 & 29.28 & 31.34 & 33.33 & 41.58 & 31.31 & 41.18 & 48.81 & 35.35 & 37.00 \\
    RoBERT-large & 33.33 & 38.29 & 37.31 & 37.50 & 27.72 & 29.29 & 36.27 & 32.14 & 40.40 & 36.69\\
    jurBERT-base & 36.56 & 33.78 & 37.31 & 35.42 & 22.77 & 42.42 & 32.35 & 35.71 & 33.33 & 34.41\\
    jurBERT-large & 44.09 & 32.43 & 47.76 & 32.81 & 35.64 & 36.36 & 36.27 & 36.90 & 39.39 & 37.96\\
     \midrule
    \textit{LLM ZS} &&&&&&&&&&\\
     FLAN-T5 XL & 39.36 & 39.29 & 41.00 & 36.08 & 40.59 & 39.39 & 40.20 & 45.24 & 39.39 & 39.44\\
     FLAN-T5 XXL & 48.94 & 40.18 & 38.00 & 37.63 & 46.53 & 37.37 & 48.04 & 52.38 & 37.37 & 42.94\\
     Mistral-Instruct v0.1 & 47.87 & 35.27 & 34.96 & 37.11 & 44.55 & 43.43 & 46.07 & 41.67 & 37.37 & 40.92\\
     Mistral-Instruct v0.2 & 41.49 & 39.73 & 35.63 & 42.27 & 44.55 & 48.48 & 49.02 & 47.62 & 37.37 & 42.80\\
     Llama-3.1 8b Instruct & 46.53 & 33.04 & 32.84 & 32.47 & 46.53 & 38.38 & 29.41 & 46.43 & 41.41 & 38.56\\
     \midrule
     \textit{LLM RAG} &&&&&&&&&& \\
     FLAN-T5 XL & 51.06 & 39.29 & 32.84 & 40.21 & 47.52 & 47.47 & 58.82 & 46.43 & 51.52 & 46.13\\
     FLAN-T5 XXL & 47.87 & 43.75 & 31.34 & 42.27 & 55.45 & 56.57 & 60.78 & 47.62 & \textbf{57.58} & 49.25 \\
     Mistral-Instruct v0.1 & 42.55 & 34.82 & 34.33 & 35.05 & 51.49 & 41.41 & 55.88 & 47.62 & 50.51 & 43.74 \\
     Mistral-Instruct v0.2 & 51.06 & 39.29 & 38.81 & 42.78 & 57.43 & 53.54 & 63.73 & 45.24 & 56.57 & 49.83 \\
     Llama-3.1 8b Instruct & \textbf{53.19} & 43.30 & 40.30 & 33.51 & 54.46 & 58.59 & 62.75 & 64.28 & \textbf{57.58} & 52.00\\
     \midrule
     \textit{LLM LFT} &&&&&&&&&& \\
     FLAN-T5 XL & 45.74 & 50.00 & 71.64 & 43.30 & 48.51 & 45.45 & 54.90 & 55.95 & 45.45 & 51.22\\
     FLAN-T5 XXL & 45.74 & 51.79 & 74.63 & 48.97 & 50.50 & 49.49 & 52.94 & \textbf{70.24} & 53.53 & 55.31 \\
     Mistral-Instruct v0.1 & 38.30 & 46.43 & 71.64 & 42.78 & 52.48 & 49.49 & 54.90 & 65.48 & 47.47 & 52.11 \\
     Mistral-Instruct v0.2 & 42.55 & 51.79 & 67.16 & 45.88 & 48.51 & 47.47 & 53.92 & \textbf{70.24} & 49.49 & 53.00\\
     Llama-3.1 8b Instruct & 45.74 & 50.00 & 68.66 & 47.94 & 54.46 & 48.48 & 51.97 & 66.67 & 55.56 & 54.39\\
     \midrule
     \textit{GRAF(Ours)} &&&&&&&&&&\\
     RoBERT-small & 35.48 & 49.10 & 67.69 & 42.11 & 50.00 & 48.96 & 48.48 & 54.79 & 45.26 & 49.10 \\
     RoBERT-base & 46.24 & 50.45 & 69.23 & 42.63 & 45.91 & \textbf{57.29} & 57.58 & 58.90 & 41.05 & 52.14\\
     RoBERT-large & 47.31 & \textbf{52.70} & 66.15 & \textbf{51.05} & \textbf{59.18} & 51.04 & 56.57 & 61.64 & 50.53 & 55.13 \\
     jurBERT-base & 49.46 & 49.55 & \textbf{78.46} & 46.84 & 53.06 & 50.00 & 57.58 & 58.90 & 48.42 & 54.70\\
     jurBERT-large & 45.16 & 47.30 & 69.23 & 45.79 & 47.96 & 54.17 & \textbf{68.69} & 67.12 & 56.84 & \textbf{55.81}\\
     \bottomrule
  \end{tabular}
  \caption{Detailed results for promotion exams.}
  \label{tab:allprom}
\end{table*}

\begin{table}
  \noindent
  \centering
  \tiny
  \begin{tabular}{lccccc}
    \toprule
    \textbf{Model} & 
    \textbf{Civil} & \textbf{Penal} & \textbf{Civil Proc.} & \textbf{Penal Proc.} &  \textbf{Average}
    \\
    \midrule
    \textit{QBERT} &&&&&\\
    RoBERT-small & 29.41 & 49.10 & 37.25 & 33.33 & 37.27 \\
    RoBERT-base & 27.45 & 45.09 & 27.45 & 25.49 & 31.37 \\
    RoBERT-large & 29.41 & 37.25 & 37.25 & 35.29 & 34.80 \\
    jurBERT-base & 41.18 & 33.33 & 33.33 & 31.37 & 34.80 \\
    jurBERT-large & 31.37 & 21.57 & 33.33 & 39.22 & 31.37 \\
    \midrule
    \textit{CrossQBERT} &&&&&\\
    RoBERT-small & 21.57 & 49.02 & 43.14 & 37.25 & 37.75 \\
    RoBERT-base & 31.37 & 27.45 & 27.45 & 25.49 & 27.94 \\
    RoBERT-large & 41.18 & 29.41 & 29.41 & 31.37 & 32.84 \\
    jurBERT-base & 25.49 & 35.29 & 35.29 & 42.14 & 34.55 \\
    jurBERT-large & 35.29 & 41.18 & 43.14 & 27.45 & 36.77 \\
    \midrule
    \textit{ColBERT} &&&&&\\
    RoBERT-small & 23.53 & 33.33 & 35.29 & 27.45 & 29.90 \\
    RoBERT-base & 33.33 & 29.41 & 27.45 & 21.57 & 27.94 \\
    RoBERT-large & 21.57 & 43.14 & 50.98 & 31.37 & 36.77 \\
    jurBERT-base & 31.37 & 33.33 & 43.14 & 41.18 & 37.26 \\
    jurBERT-large & 45.10 & 41.18 & 29.41 & 31.37 & 36.77 \\
     \midrule
     \textit{LLM ZS} &&&&&\\
     FLAN-T5 XL & 29.41 & 21.57 & 13.72 & 27.45 & 23.04 \\
     FLAN-T5 XXL & 31.37 & 25.49 & 15.69 & 25.49 & 24.51 \\
     Mistral-Instruct v0.1 & 33.33 & 21.57 & 9.80 & 25.49 & 22.55 \\
     Mistral-Instruct v0.2 & 39.21 & 19.61 & 11.76 & 21.57 & 23.04 \\
     Llama-3.1 8b Instruct & 33.33 & 31.37 & 17.65 & 27.46 & 27.45 \\
     \midrule
     \textit{LLM RAG} &&&&& \\
     FLAN-T5 XL & 31.37 & 17.65 & 21.57 & 23.53 & 23.53 \\
     FLAN-T5 XXL & 39.22 & 27.45 & 21.57 & 27.45 & 28.92 \\
     Mistral-Instruct v0.1 & 11.76 & 7.84 & 5.88 & 15.69 & 10.29 \\
     Mistral-Instruct v0.2 & 43.14 & 23.53 & 21.57 & 27.45 & 28.92 \\
     Llama-3.1 8b Instruct & 35.29 & 11.76 & 23.53 & 13.73 & 21.08 \\
     \midrule
     \textit{LLM LFT} &&&&& \\
     FLAN-T5 XL & 35.29 & 45.10 & 41.18 & 41.18 & 40.69 \\
     FLAN-T5 XXL & 37.25 & 43.14 & 35.29 & 29.41 & 36.27 \\
     Mistral-Instruct v0.1 & 19.61 & 29.41 & 29.41 & 27.45 & 26.47 \\
     Mistral-Instruct v0.2 & 17.65 & 17.65 & 17.65 & 23.53 & 19.12 \\
     Llama-3.1 8b Instruct & 31.76 & 33.33 & 19.61 & 35.29 & 30.00 \\
     \midrule
     \textit{GRAF(Ours)} &&&&&\\
     RoBERT-small & 43.14 & 52.94 & 45.10 & 45.10 & 46.57 \\
     RoBERT-base & \textbf{60.78} & 58.82 & 41.18 & 50.98 & 52.94 \\
     RoBERT-large & 52.94 & \textbf{62.75} & \textbf{54.90} & \textbf{56.86} & \textbf{56.86} \\
     jurBERT-base & 47.06 & 52.94 & 50.98 & 45.10 & 49.02 \\
     jurBERT-large & 54.90 & 54.90 & 52.94 & 49.01 & 52.94  \\
     \bottomrule
  \end{tabular}
  \caption{Detailed results for entrance exams.}
  \label{tab:allentrance}
\end{table}

\begin{table}
  \tiny
  \centering
  \begin{tabular}{lccccc}
    \toprule
    \textbf{Model} & 
    \textbf{Civil} & \textbf{Penal} & \textbf{Civil Proc.} & \textbf{Penal Proc.} &  \textbf{Average}
    \\
    \midrule
    \textit{QBERT} &&&&&\\
    RoBERT-small & 37.90 & 32.26 & 33.06 & 32.26 & 33.87 \\
    RoBERT-base & 26.61 & 27.42 & 29.84 & 36.29 & 30.04 \\
    RoBERT-large & 34.68 & 32.26 & 40.32 & 33.87 & 35.28 \\
    jurBERT-base & 33.87 & 23.58 & 33.06 & 33.06 & 30.89 \\
    jurBERT-large & 40.32 & 31.71 & 32.26 & 30.65 & 33.74 \\
    \midrule
    \textit{CrossQBERT} &&&&&\\
    RoBERT-small & 34.68 & 31.97 & 39.52 & 34.68 & 35.21 \\
    RoBERT-base & 30.65 & 38.52 & 29.84 & 30.65 & 32.42 \\
    RoBERT-large & 41.94 & 36.07 & 33.87 & 34.68 & 36.64 \\
    jurBERT-base & 22.58 & 34.43 & 29.84 & 34.68 & 30.28 \\
    jurBERT-large & 36.29 & 27.05 & 31.45 & 30.65 & 31.36 \\
    \midrule
    \textit{ColBERT} &&&&&\\
    RoBERT-small & 29.03 & 33.06 & 28.23 & 27.42 & 29.44 \\
    RoBERT-base & 38.71 & 27.42 & 34.68 & 29.84 & 32.66 \\
    RoBERT-large & 36.29 & 32.26 & 36.29 & 39.52 & 36.09 \\
    jurBERT-base & 23.39 & 29.03 & 31.45 & 35.48 & 29.84 \\
    jurBERT-large & 29.03 & 33.87 & 35.48 & 33.87 & 33.06 \\
     \midrule
     \textit{LLM ZS} &&&&&\\
     FLAN-T5 XL & 19.35 & 25.00 & 17.74 & 12.90 & 18.75 \\
     FLAN-T5 XXL & 21.77 & 24.19 & 23.38 & 18.54 & 21.97 \\
     Mistral-Instruct v0.1 & 14.52 & 12.10 & 16.13 & 13.70 & 14.11 \\
     Mistral-Instruct v0.2 & 19.35 & 22.58 & 15.32 & 14.52 & 17.94 \\
     Llama-3.1 8b Instruct & 16.13 & 24.19 & 20.16 & 19.35 & 19.96 \\
     \midrule
     \textit{LLM RAG} &&&&& \\
     FLAN-T5 XL & 21.77 & 22.58 & 20.16 & 16.94 & 20.36 \\
     FLAN-T5 XXL & 20.97 & 26.61 & 33.87 & 20.97 & 25.61 \\
     Mistral-Instruct v0.1 & 5.65 & 11.29 & 12.90 & 5.65 & 8.87 \\
     Mistral-Instruct v0.2 & 26.61 & 29.03 & 20.16 & 19.35 & 23.79 \\
     Llama-3.1 8b Instruct & 25.00 & 26.61 & 16.12 & 23.38 & 22.78 \\
     \midrule
     \textit{LLM LFT} &&&&& \\
     FLAN-T5 XL & 31.45 & 32.26 & 29.03 & 36.29 & 32.26 \\
     FLAN-T5 XXL & 27.42 & 28.23 & 33.06 & 22.58 & 27.82 \\
     Mistral-Instruct v0.1 & 27.42 & 31.45 & 21.77 & 29.03 & 27.42 \\
     Mistral-Instruct v0.2 & 13.71 & 14.52 & 24.19 & 23.39 & 18.95 \\
     Llama-3.1 8b Instruct & 25.80 & 24.19 & 25.00 & 23.39 & 24.60 \\
     \midrule
     \textit{GRAF(Ours)} &&&&&\\
     RoBERT-small & 46.77 & 39.52 & 43.90 & 43.44 & 43.41 \\
     RoBERT-base & 48.39 & 49.19 & 52.85 & 46.72 & 49.29 \\
     RoBERT-large & 50.00 & 47.58 & 49.59 & 55.73 & 50.73 \\
     jurBERT-base & \textbf{51.61} & 55.65 & 51.22 & \textbf{59.02} & 54.38 \\
     jurBERT-large & \textbf{51.61} & \textbf{61.29} & \textbf{56.10} & 57.38 & \textbf{56.60}  \\
     \bottomrule
  \end{tabular}
  \caption{Detailed results for bar exams.}
  \label{tab:allbar}
\end{table}

\clearpage

\section{Romanian Prompts}
\label{appendix:ro_prompts}

\prompt{18em}{Prompt 1 -- Entrance and Bar Exams}
{
Răspunde la următoarea întrebare de legalitate din \{tip drept\}. Cel mult 2 răspunsuri sunt corecte.

Dacă un singur răspuns este corect, vei răspunde doar cu litera răspunsului corect.

Dacă 2 răspunsuri sunt corecte, vei răspunde doar cu literele răspunsurilor corecte:\\

\{tip drept\}

\{întrebare\}

\{variante de răspuns\}
}

\prompt{18em}{Prompt 1 -- Promotion Exam}
{
Răspunde la următoarea întrebare de legalitate din \{tip drept\}. Un singur răspuns este corect. Tu vei răspunde doar cu litera răspunsului corect: \\

\{tip drept\}

\{întrebare\}

\{variante de răspuns\}
}

\prompt{18em}{Prompt 2 -- Entrance and Bar Exams}
{
Răspunde la următoarea întrebare de legalitate din \{tip drept\}. Cel mult 2 răspunsuri sunt corecte.

Dacă un singur răspuns este corect, vei răspunde doar cu litera răspunsului corect.

Dacă 2 răspunsuri sunt corecte, vei răspunde doar cu literele răspunsurilor corecte

Răspunde cu doar unul dintre simbolurile din lista [A, B, C, AB, AC, BC]:\\

\{tip drept\}

\{întrebare\}

\{variante de răspuns\}
}

\prompt{18em}{Prompt 2 -- Promotion Exam}
{
Răspunde la următoarea întrebare de legalitate din \{tip drept\} cu doar una dintre literele din lista [A, B, C]. Un singur răspuns este corect: \\

\{tip drept\}

\{întrebare\}

\{variante de răspuns\}
}

\prompt{18em}{FLAN-T5 RAG -- Entrance and Bar Exams}
{
Răspunde la următoarea întrebare de legalitate din \{tip drept\} din context. Cel mult 2 răspunsuri sunt corecte.

Dacă un singur răspuns este corect, vei răspunde doar cu litera răspunsului corect.

Dacă 2 răspunsuri sunt corecte, vei răspunde doar cu literele răspunsurilor corecte

Dacă informația din context nu este în întrebare atunci ignoră contextul și doar răspunde la întrebare.

Răspunde cu doar unul dintre simbolurile din lista [A, B, C, AB, AC, BC]. \\

Context:

\{documente\} \\

Întrebare:

\{întrebare\}

\{variante de răspuns\}
}

\prompt{18em}{FLAN-T5 RAG -- Promotion Exams}
{
Răspunde la următoarea întrebare de legalitate din \{tip drept\} din context cu doar una dintre literele din lista [A, B, C].

Dacă informația din context nu este în întrebare atunci ignoră contextul și doar răspunde la întrebare.

Un singur răspuns este corect. \\

Context:

\{documente\} \\

Întrebare:

\{întrebare\}

\{variante de răspuns\}
}

\prompt{18em}{Mistral RAG -- Entrance and Bar Exams}
{
Răspunde la următoarea întrebare de legalitate din \{tip drept\} din context. Cel mult 2 răspunsuri sunt corecte.

Dacă un singur răspuns este corect, vei răspunde doar cu litera răspunsului corect.

Dacă 2 răspunsuri sunt corecte, vei răspunde doar cu literele răspunsurilor corecte.

Dacă informația din context nu este în întrebare atunci ignoră contextul și doar răspunde la întrebare. \\

Context:

\{documente\} \\

Întrebare:

\{întrebare\}

\{variante de răspuns\}
}

\prompt{18em}{Mistral RAG -- Promotion Exam}
{
Răspunde la următoarea întrebare de legalitate din \{tip drept\} din context. Un singur răspuns este corect. Tu vei răspunde doar cu litera răspunsului corect.

Dacă informația din context nu este în întrebare atunci ignoră contextul și doar răspunde la întrebare. \\

Context:

\{documente\} \\

Întrebare:

\{întrebare\}

\{variante de răspuns\}
}

\promptLarge{40em}{LLM Prompt for Claim Graph Extraction}
{
Extrage toate entitățile și toate relațiile dintre entități din textul legal pe baza exemplului. La final adaugă STOP.
Tu vei răspunde cu triplete de forma: (entitate;relație;entitate). Tripletele sunt separate pe linii. Fiecare relație triplet se va trece separat.
Entitățile pot fi instituții, organizații, persoane, funcții, documente, instanțe și altele.\\

Text: \\
(1) Pe lângă fiecare curte de apel va funcţiona o comisie de cercetare a averilor, denumită în continuare comisie de cercetare, formată din:\\
a) 2 judecători de la curtea de apel, desemnaţi de preşedintele acesteia, dintre care unul în calitate de preşedinte,\\
b) un procuror de la parchetul care funcţionează pe lângă curtea de apel, desemnat de prim-procurorul acestui parchet.\\
(2) Preşedintele şi membrii comisiei de cercetare sunt desemnaţi pe o perioadă de 3 ani. Pe aceeaşi perioadă şi de către aceleaşi persoane vor fi desemnaţi şi 3 supleanţi, care îi vor înlocui pe titulari în cazul în care aceştia, din motive legale, nu vor putea lua parte la lucrările comisiei de cercetare.\\
(3) Comisia de cercetare are un secretar, desemnat de preşedintele curţii de apel dintre grefierii acestei instanţe.\\
 
Entitate;Relație;Entitate:\\
(curte de apel;funcționează pe lângă;comisie de cercetare a averilor)\\
(comisie de cercetare a averilor;denumită;comisie de cercetare)\\
(comisie de cercetare;formată din;2 judecători)\\
(2 judecători;desemnați de;președinte curte de apel)\\
(comisie de cercetare;formată din;procuror)\\
(procuror;de la;parchetul care funcționează pe lângă curtea de apel)\\
(procuror;desemnat de;prim-procuror)\\
(președinte comisie de cercetare;desemnat pe o perioadă de;3 ani)\\
(membrii comisiei de cercetare;desemnat pe o perioadă de;3 ani)\\
(3 supleanți;desemnați de;președinte curte de apel)\\
(3 supleanți;desemnați de;prim-procuror)\\
(3 supleanți;desemnați pe o perioadă de;3 ani)\\
(3 supleanți;îi vor înlocui dacă nu vor putea lua parte la lucrările comisiei de cercetare pe;titulari)\\
(comisie de cercetare;are;un secretar)\\
(un secretar;desemnat dintre grefieri de;președinte curte de apel)\\
STOP\\
 
Text:\\
\{text\}\\
 
Entitate;Relație;Entitate:
}

\clearpage

\section{Translated Prompts}
\label{appendix:en_prompts}

\prompt{18em}{Prompt 1 -- Entrance and Bar Exams}
{
Answer the following legal question from \{law type\}. At most 2 answers are correct.

If a single answer is correct, you will only answer with the letter of the correct answer.

If 2 answers are correct, you will answer only with the letters of the correct answers:\\

\{law type\}

\{question\}

\{answer choices\}
}

\prompt{18em}{Prompt 1 -- Promotion Exam}
{
Answer the following legal question from \{law type\}. A single answer is correct. You will only answer with the letter of the correct answer: \\

\{law type\}

\{question\}

\{answer choices\}
}

\prompt{18em}{Prompt 2 -- Entrance and Bar Exams}
{
Answer the following question from \{law type\}. At most 2 answers are correct.

If a single answer is correct, you will only have an answer with the letter of the correct answer.

If 2 answers are correct, you will answer with the letters of the correct letters.

Answer with only one of the symbols from the list [A, B, C, AB, AC, BC]:\\

\{law type\}

\{question\}

\{answer choices\}
}

\prompt{18em}{Prompt 2 -- Promotion Exam}
{
Answer the following legal question from \{law type\} with only one of the letters from the list [A, B, C]. A single answer is correct: \\

\{law type\}

\{question\}

\{answer choices\}
}

\prompt{18em}{FLAN-T5 RAG -- Entrance and Bar Exams}
{
Answer the following legal question from \{law type\}. At most 2 answers are correct.

If a single answer is correct, you will only answer with the letter of the correct answer.

If 2 answers are correct, you will answer with only the letters of the correct answers.

If the information from the context is not in the question, then ignore the context and answer the question.

Answer with only one of the symbols from the list [A, B, C, AB, AC, BC]. \\

Context:

\{documets\} \\

Question:

\{question\}

\{answer choices\}
}

\prompt{18em}{FLAN-T5 RAG -- Promotion Exam}
{
Answer the following legal question from \{law type\} with only one of the letters from the list [A, B, C].

If the information from the context is not in the question, then ignore the context and answer the question.

A single answer is correct. \\

Context:

\{documents\} \\

Question:

\{question\}

\{answer choices\}
}

\prompt{18em}{Mistral RAG -- Entrance and Bar Exams}
{
Answer the following legal question from \{law type\}. At most 2 answers are correct.

If a single answer is correct, you will only answer with the letter of the correct answer.

If 2 answers are correct, you will answer only with the letters of the correct answers.

If the information from the context is not in the question, then ignore the context and answer the question. \\

Context:

\{documents\} \\

Question:

\{question\}

\{answer choices\}
}

\prompt{18em}{Mistral RAG -- Promotion Exam}
{
Answer the following legal question from \{law type\}. You will only answer with the letter of the correct answer.

If the information from the context is not in the question, then ignore the context and answer the question.

A single answer is correct. \\

Context:

\{documents\} \\

Question:

\{question\}

\{answer choices\}
}

\promptLarge{40em}{LLM Prompt for Claim Graph Extraction}
{
Extract all entities and relationships between entities from the legal text based on the example. In the end, add STOP.
You will answer with triplets of the form: (entity;relation;entity). The triplets are separated on lines. Each triplet relationship will be entered separately.
Entities can be institutions, organizations, persons, functions, documents, courts and others.\\

Text:\\
(1) An assets investigation commission, hereinafter referred to as the investigation commission, shall operate in addition to each court of appeal, consisting of:\\
a) 2 judges from the court of appeal, designated by its president, one of whom shall act as president,\\
b) a prosecutor from the prosecutor's office operating under the court of appeal, designated by the chief prosecutor of this prosecutor's office.\\
(2) The president and members of the investigation commission shall be designated for a period of 3 years. During the same period and by the same persons, 3 alternates will also be appointed, who will replace the holders in the event that they, for legal reasons, are unable to participate in the work of the investigation commission.\\
(3) The investigation commission has a secretary, appointed by the president of the court of appeal from among the clerks of this court.\\

Entity;Relationship;Entity:\\
(court of appeal;shall operated in addition to;assets investigation commission)\\
(assets investigation commission;referred to as;investigation commission)\\
(investigation commission;consisting of;2 judges)\\
(2 judges;designated by;president of the court of appeal)\\
(investigation commission;consisting of;prosecutor)\\
(prosecutor;from;prosecutor's office operating under the court of appeal)\\
(prosecutor;designated by;chief prosecutor)\\
(president of the investigation commission;designated for a period of;3 years)\\
(members of the investigation commission;designated for a period of;3 years)\\
(3 alternates;appointed by;the president of the court of appeal)\\
(3 alternates;appointed by;the chief prosecutor)\\
(3 alternates;designated for a period of;3 years)\\
(3 alternates;will replace the holders if they cannot take part in the work of the investigation commission on;the heads)\\
(investigation commission;has;a secretary)\\
(a secretary;appointed by among the clerks of;the president of the court of appeal)\\
STOP\\

Text:\\
\{text\}\\

Entity;Relationship;Entity:
}

\clearpage
\onecolumn
\section{Figures and Tables for Topic Analysis}

\begin{table*}[!bh]
\small
\begin{tabular}{p{5cm}p{10cm}}
\toprule
\textbf{Topic} & \textbf{Keywords} \\
\midrule
Apel și Instanță & apel, în, poate, de, care, nu, judecată, instanţa, fi, se \\
\hline
Debitor și Creditor & debitorului, debitorul, debitor, creditor, creditorul, creditorilor, creditorului, insolvenţă, procedurii, insolvenţei \\
\hline
Omor și Victimă & omor, faptei, rezultatul, infracţiunea, victimei, fapta, autorul, făptuitorul, culpă, victima \\
\hline
Proprietate și Contracte & bunului, dreptul, vânzătorul, bunul, vânzării, proprietate, vânzare, cumpărătorului, contractului, cumpărătorul \\
\hline
Termene și Comunicare & zile, termen, data, comunicare, la, termenul, de, comunicării, comunicarea, 30 \\
\hline
Pedeapsă și Condamnare & pedepsei, pedeapsa, vigoare, supraveghere, pedeapsă, amenzii, închisorii, condamnare, legea, executării \\
\hline
România și Sistemul Legal & româniei, românia, român, română, arestare, române, european, decizia, teritoriul, monitorul \\
\hline
Căsătorie și Divorț & căsătoriei, divorţ, soţi, căsătoria, căsătorie, divorţul, desfacerea, culpa, casarea, legea \\
\hline
Furt și Răspundere Penală & infracţiunea, furt, libertate, tentativa, infracțiunea, lipsire, posibilă, art, infracțiunile, infracţiunii \\
\hline
Copil și Drepturi Părintești & copilului, părinteşti, copilul, minorului, părinţilor, vârsta, copil, părintele, drepturile, protecţie \\
\hline
Contracte și Obligații & contractului, contractul, încheiat, contract, mandatarul, mandatului, mandantului, produce, secret, este \\
\hline
Litigii și Părți & bb, aa, reclamantul, judecată, pârâtul, chemare, lei, cerere, acest, contradictoriu \\
\hline
Moștenire și Succesiune & defunctului, moștenire, moştenire, succesorală, moștenirea, moştenirii, lui, moștenirii, culege, moştenirea \\
\hline
Pensie și Retragere din Activitate & cotizare, stagiul, pensiei, invaliditate, pensii, pensie, standard, realizat, pensionare, pensia \\
\hline
Sentință Penală și Apeluri & pedeapsa, ani, închisoare, inculpatul, închisorii, pedepsei, reabilitare, apel, apelul, condamnat \\
\hline
Fals și Fraudă & fals, înscrisul, privată, semnătură, înscrisuri, înscrisului, sub, înscris, 250, oficiale \\
\hline
Drept Administrativ & administrativ, administrative, contencios, publice, actul, act, actului, actele, publică, nelegalitate \\
\hline
Suspendare Judiciară și Hotărâri & suspendarea, suspendare, justiţie, apel, hotărâri, judecăţii, se, rolul, primă, dispusă \\
\hline
Constituționalitate și Lege & art, alin, neconstituţionalitate, curtea, prevederile, constituţională, din, constituţionale, că, excepţia \\
\hline
Mită și Corupție & serviciu, mită, luare, bani, public, abuz, funcţionar, foloase, infracţiunii, sumă \\
\hline
Drept Comercial și Acționari & societăţii, adunarea, social, adunării, acţiuni, generală, societate, capitalul, generale, vot \\
\hline
Furt Calificat și Acuzații & lui, infracțiunea, furt, și, calificat, pe, sarcina, că, inculpatul, un \\
\hline
Competență și Jurisdicție a Instanțelor & grad, conexitate, instanţe, cereri, litispendenţa, divergență, litispendenţă, cealaltă, invocată, două \\
\hline
Paternitate și Parentalitate & copilului, mamei, paternităţii, paternitate, copilul, tată, acţiunea, căsătoriei, născut, timpul \\
\hline
Reglementări UE și Conformitate & membru, stat, statul, regulamentului, european, executoriu, 44, nr, 2001, materie \\
\hline
Competență Legală și Conflicte de Jurisdicție & competenţa, competenţă, apel, instanţei, conflictul, instanţe, competente, rediscutată, îşi, dintre \\
\hline
Citație și Chemare în Instanță & prezentă, partea, termenul, citare, citată, termen, amânarea, studierii, legal, fost \\
\hline
Excepții Procedurale și Erori & procesuale, capacităţii, excepţia, lipsei, folosinţă, excepţiile, invocate, active, erori, necompetenţa \\
\hline
Infracțiuni Sexuale și Viol & viol, sexual, incest, agravată, violare, infracţiunea, sexuală, infracțiunea, domiciliu, prevăzută \\
\hline
Instanțele din București și Jurisdicție & bucurești, ab, judecătoria, in, procurorul, sector, disp, lângă, pen, suspectul \\
\bottomrule
\end{tabular}
\caption{List of top 30 topics and associated keywords from the JuRO dataset, in Romanian.}
\label{tab:top_30_juro_ro}
\end{table*}

\begin{table*}
\small
\begin{tabular}{p{5cm}p{10cm}}
\toprule
\textbf{Topic} & \textbf{Keywords} \\ 
\midrule
Appeal and Court & appeal, in, may, of, which, not, trial, court, be, is \\
\hline
Debtor and Creditor & debtor's, debtor, debtor, creditor, creditor's, creditors, creditor's, insolvency, procedure, insolvency \\
\hline
Homicide and Victim & homicide, act, result, crime, victim's, act, author, perpetrator, guilt, victim \\
\hline
Property and Contracts & asset's, right, seller, asset, sale, property, sale, buyer's, contract's, buyer \\
\hline
Deadlines and Communication & days, deadline, date, communication, at, term, of, communication's, communication, 30 \\
\hline
Punishment and Sentencing & penalty's, penalty, enforcement, supervision, punishment, fine, imprisonment, conviction, law, execution \\
\hline
Romania and Legal System & Romania's, Romania, Romanian, Romanian, arrest, Romanian, European, decision, territory, official journal \\
\hline
Marriage and Divorce & marriage's, divorce, spouses, marriage, marriage, divorce, dissolution, fault, annulment, law \\
\hline
Theft and Criminal Liability & crime, theft, freedom, attempt, crime, deprivation, possible, article, crimes, crime \\
\hline
Child and Parental Rights & child's, parental, child, minor's, parents', age, child, parent, rights, protection \\
\hline
Contracts and Obligations & contract's, contract, concluded, contract, agent, mandate's, principal's, produce, secret, is \\
\hline
Litigation and Parties & bb, aa, claimant, trial, defendant, summons, lei, request, this, adversarial \\
\hline
Inheritance and Succession & deceased's, inheritance, inheritance, succession, inheritance, inheritance's, his, inheritance's, collects, inheritance \\
\hline
Pension and Retirement & contribution, period, pension's, disability, pensions, pension, standard, achieved, retirement, pension \\
\hline
Criminal Sentencing and Appeals & penalty, years, imprisonment, defendant, imprisonment, penalty's, rehabilitation, appeal, appeal, convicted \\
\hline
Forgery and Fraud & forgery, document, private, signature, documents, document's, under, document, 250, official \\
\hline
Administrative Law & administrative, administrative, litigation, public, act, act, act's, acts, public, illegality \\
\hline
Judicial Suspension and Rulings & suspension, suspension, justice, appeal, rulings, court, is, role, first, ordered \\
\hline
Constitutionality and Law & article, paragraph, unconstitutionality, court, provisions, constitutional, from, constitutional, that, exception \\
\hline
Bribery and Corruption & service, bribe, taking, money, public, abuse, official, benefits, crime, sum \\
\hline
Corporate Law and Shareholders & company's, assembly, social, assembly's, shares, general, company, capital, general, vote \\
\hline
Qualified Theft and Charges & his, crime, theft, and, qualified, on, charge, that, defendant, a \\
\hline
Jurisdiction and Court Competence & level, connection, courts, requests, lis pendens, divergence, lis pendens, other, invoked, two \\
\hline
Paternity and Parenthood & child's, mother's, paternity's, paternity, child, father, action, marriage's, born, time \\
\hline
EU Regulations and Compliance & member, state, state's, regulation's, European, enforceable, 44, no, 2001, matter \\
\hline
Legal Competence and Jurisdiction Conflicts & competence, competence, appeal, court's, conflict, courts, competent, reconsidered, its, among \\
\hline
Court Summons and Citations & present, party, deadline, citation, cited, term, postponement, study, legal, was \\
\hline
Procedural Exceptions and Errors & procedural, capacity's, exception, lack, use, exceptions, invoked, active, errors, incompetence \\
\hline
Sexual Crimes and Violations & rape, sexual, incest, aggravated, violation, crime, sexual, crime, domicile, provided \\
\hline
Bucharest Courts and Jurisdiction & Bucharest, ab, court, in, prosecutor, sector, ordered, near, criminal, suspect \\
\bottomrule
\end{tabular}
\caption{List of top 30 topics and associated keywords from the JuRO dataset, translated to English.}
\label{tab:top_30_juro_en}
\end{table*}

\begin{table*}
\small
\begin{tabular}{p{5cm}p{10cm}}
\toprule
\textbf{Topic} & \textbf{Keywords} \\
\midrule
Administrație Publică și Guvernare & publice, publici, de, al, şi, sau, în, consiliului, și, care \\
\hline
România și Afaceri Europene & românia, româniei, în, de, din, europene, române, și, la, al \\
\hline
Protecția Copilului și Dreptul Familiei & copilului, copilul, tutelă, minorului, familie, copil, vârsta, minorul, de, tutore \\
\hline
Reglementări Legale și Conformitate & articolul, următorul, cuprins, avea, modifică, alineatul, va, şi, la, se \\
\hline
Impozitare și Politici Fiscale & fiscal, fiscale, fiscală, din, şi, de, pentru, organul, al, în \\
\hline
Căsătorie, Divorț și Drept Matrimonial & căsătoriei, căsătoria, soţi, căsătorie, soți, matrimonial, divorț, dintre, comune, soții \\
\hline
Pedepse Legale și Sancțiuni & ani, amendă, închisoare, închisoarea, 000, pedepseşte, lei, pedeapsa, la, cu \\
\hline
Moștenire și Dreptul Succesiunii & moştenire, moștenire, moştenirii, moștenirii, defunctului, moştenirea, moștenirea, privilegiaţi, succesorală, privilegiați \\
\hline
Transport și Drept Maritim & transport, transportatorul, vasului, transportului, transportatorului, vasul, expeditorul, capitanul, sau, pentru \\
\hline
Deteriorarea Proprietății și Răspundere Legală & bunului, prejudiciul, locatarul, prejudiciului, cauzat, dacă, bunul, este, repararea, să \\
\hline
Dreptul Muncii și Drepturile Angajaților & muncă, colective, colectiv, angajatorul, sindicale, individual, salariatul, salariatului, muncii, de \\
\hline
Servicii Digitale și Dreptul Informației & publice, servicii, electronic, electronice, și, ministerul, furnizorul, electronică, şi, informației \\
\hline
Termene Legale și Expirare & termenul, zi, termen, ziua, ani, prescrie, zile, ultima, data, termenului \\
\hline
Reforme Legale și Amendamente & 2019, publicată, monitorul, oficial, 597, ix, 07, 05, ordonanța, urgență \\
\hline
Securitate Socială și Dreptul Pensiilor & investiţii, şi, cnpp, snpct, pensii, 01, de, consiliul, din, financiare \\
\hline
Certificate de Deces și Declarații Legale & decedate, moartea, morţii, morții, decesul, mort, cel, declarat, viaţă, viață \\
\hline
Proprietate și Posesiune & mobile, bunurile, mobil, bunurilor, bun, bunuri, bunului, imobile, debitorului, sunt \\
\hline
Legislație de Urgență și Ordonanțe & ordonanţa, urgenţă, din, nan, ordonanța, urgență, persoanele, astăzi, stat, de \\
\hline
Contracte și Acorduri Legale & persoanei, private, unei, fără, utilizarea, sau, difuzarea, acordul, persoane, precum \\
\hline
Fructe și Drept Agricol & fructele, fructelor, industriale, naturale, cuvin, bunului, recoltelor, civile, fructe, dreptul \\
\hline
Proprietate Funciară și Cadastru & zidului, comune, zidul, linia, hotar, clădirii, apartamente, clădire, necomunitate, proprietari \\
\hline
Drepturi de Proprietate și Liberalități & liberalitatea, liberalității, liberalităţii, dispunătorul, instituitului, impută, primi, dispunător, rezervatari, liberalitate \\
\hline
Drepturile Omului și Identitate Civilă & umane, fizică, civile, imagine, portret, psihică, dreptul, ființei, orice, civilă \\
\hline
Identificare Legală și Aspecte Familiale & identificare, numele, numelui, nume, persoanei, prenumele, atribute, juridice, familie, identificarea \\
\hline
Transfer Juridic și Divizare & juridice, persoanei, organic, divizarea, patrimoniul, multe, legea, juridică, persoane, transferă \\
\hline
Dreptul Ospitalității și Răspundere & hotelierul, hotel, hotelierului, clientului, aduse, cazare, hotelului, bunurilor, cazării, răspunderea \\
\hline
Capacitate Juridică și Drepturile Minorilor & exerciţiu, exercițiu, capacitate, capacitatea, restrânsă, deplină, minorul, actele, lipsit, tutelă \\
\hline
Solidaritate și Răspundere în Grupuri & solidar, răspund, nelimitat, primii, fondatorii, grupului, prejudiciul, fondatori, cauzat, sînt \\
\hline
Datorii și Drepturile Creditorilor & cedată, creanța, creanţa, creanței, cesiune, cesiunea, creanţei, cesionarul, cesionarului, constatator \\
\hline
Instrumente Financiare și Dreptul Garanției & girul, alb, gir, cec, litere, suma, carat, sa, avalul, este \\
\bottomrule
\end{tabular}
\caption{List of top 30 topics and associated keywords from the CROL dataset, in Romanian.}
\label{tab:top_30_crol_ro}
\end{table*}

\begin{table*}
\small
\begin{tabular}{p{5cm}p{10cm}}
\toprule
\textbf{Topic} & \textbf{Keywords} \\
\midrule
Public Administration and Governance & public, public, of, the, and, or, in, council, and, which \\
\hline
Romania and European Affairs & romania, romania's, in, of, from, european, romanian, and, at, the \\
\hline
Child Protection and Family Law & child, the child, guardianship, minor, family, child, age, the minor, of, tutor \\
\hline
Legal Regulations and Compliance & article, following, content, have, modifies, paragraph, will, and, at, is \\
\hline
Taxation and Fiscal Policies & fiscal, fiscal, fiscal, from, and, of, for, body, the, in \\
\hline
Marriage, Divorce, and Matrimonial Law & marriage, the marriage, spouses, marriage, spouses, matrimonial, divorce, between, common, spouses \\
\hline
Legal Penalties and Sanctions & years, fine, imprisonment, the imprisonment, 000, punishes, lei, penalty, at, with \\
\hline
Inheritance and Succession Law & inheritance, inheritance, inheritance, inheritance, of the deceased, the inheritance, inheritance, privileged, succession, privileged \\
\hline
Transport and Maritime Law & transport, transporter, vessel, transport, the transporter, vessel, sender, captain, or, for \\
\hline
Property Damage and Liability Law & property, damage, tenant, the damage, caused, if, the property, is, repair, to \\
\hline
Labor Law and Employment Rights & work, collective, collective, employer, union, individual, employee, the employee, work, of \\
\hline
Digital Services and Information Law & public, services, electronic, electronic, and, ministry, provider, electronic, and, information \\
\hline
Legal Timeframes and Expiration & term, day, term, the day, years, prescribes, days, last, date, term \\
\hline
Legal Reforms and Amendments & 2019, published, monitor, official, 597, ix, 07, 05, ordinance, emergency \\
\hline
Social Security and Pension Law & investments, and, cnpp, snpct, pensions, 01, of, council, from, financial \\
\hline
Death Certificates and Legal Declarations & deceased, death, of death, of death, death, dead, the, declared, life, life \\
\hline
Property Ownership and Possession & movable, properties, mobile, properties, property, goods, property, real estate, debtor, are \\
\hline
Emergency Legislation and Ordinances & ordinance, emergency, from, nan, ordinance, emergency, people, today, state, of \\
\hline
Contracts and Legal Agreements & person, private, one, without, use, or, dissemination, agreement, persons, such as \\
\hline
Fruits and Agricultural Law & fruits, of the fruits, industrial, natural, words, property, harvests, civil, fruit, right \\
\hline
Land Ownership and Cadastre & wall, common, the wall, line, border, building, apartments, building, non-community, owners \\
\hline
Property Rights and Liberal Ownership & liberality, liberality, liberality, disposer, instituted, imputes, receive, disposer, reserved, liberality \\
\hline
Human Rights and Civil Identity & human, physical, civil, image, portrait, mental, right, being, any, civil \\
\hline
Legal Identification and Family Matters & identification, name, the name, name, person, first name, attributes, legal, family, identification \\
\hline
Juridical Transfer and Division & legal, person, organic, division, patrimony, many, law, legal, persons, transfers \\
\hline
Hospitality Law and Liability & hotelier, hotel, the hotelier, client, damages, accommodation, hotel, goods, accommodation, liability \\
\hline
Legal Capacity and Minor Rights & exercise, exercise, capacity, the capacity, restricted, full, the minor, acts, deprived, guardianship \\
\hline
Solidarity and Liability in Groups & solidarity, respond, unlimited, first, founders, group, damage, founders, caused, are \\
\hline
Debt and Creditors' Rights & assigned, claim, claim, claim, assignment, assignment, claim, assignee, assignee, certifier \\
\hline
Financial Instruments and Surety Law & endorsement, blank, endorsement, check, letters, amount, carat, his, guarantee, is \\
\bottomrule
\end{tabular}
\caption{List of top 30 topics and associated keywords from the CROL dataset, translated to English.}
\label{tab:top_30_crol_en}
\end{table*}

\end{document}